\documentclass[iicol]{sn-jnl}  

\usepackage{graphicx}%
\usepackage{multirow}%
\usepackage{amsmath,amssymb,amsfonts}%
\usepackage{amsthm}%
\usepackage{mathrsfs}%
\usepackage[title]{appendix}%
\usepackage{xcolor}%
\usepackage{textcomp}%
\usepackage{manyfoot}%
\usepackage{booktabs}%
\usepackage{listings}%

\usepackage[linesnumbered,ruled]{algorithm2e}
\usepackage{algpseudocode}%

\usepackage{subcaption}
\usepackage{color}
\usepackage{pifont}  
\newcommand{\tabincell}[2]{\begin{tabular}{@{}#1@{}}#2\end{tabular}}

\theoremstyle{thmstyleone}%
\theoremstyle{thmstyletwo}%

\theoremstyle{thmstylethree}%

\begin{document}

\title[Article Title]{Exploring Decoupled Spatio-Temporal Consistency Learning and Self-Prompting Evolution for Self-Supervised Tracking}

\def\tracker{SSTrack++}


\author[1,2]{\fnm{Yaozong} \sur{Zheng}}\email{yaozongzheng@stu.gxnu.edu.cn}

\author*[1,2]{\fnm{Bineng} \sur{Zhong}}\email{bnzhong@gxnu.edu.cn}

\author[1,2]{\fnm{Qihua} \sur{Liang}}\email{qhliang@gxnu.edu.cn}

\author[1,2]{\fnm{Ning} \sur{Li}}\email{ningli65536@mailbox.gxnu.edu.cn}

\author[1,2]{\fnm{Haiying} \sur{Xia}}\email{xhy22@mailbox.gxnu.edu.cn}

\author[1,2]{\fnm{Shuxiang} \sur{Song}}\email{songshuxiang@mailbox.gxnu.edu.cn}

\author[3]{\fnm{Rongrong} \sur{Ji}}\email{rrji@xmu.edu.cn}

\affil[1]{\orgdiv{Key Laboratory of Education Blockchain and Intelligent Technology, Ministry of Education}, \orgname{Guangxi Normal University}, \orgaddress{\city{Guilin}, \postcode{541004}, \state{Guangxi Province}, \country{China}}}

\affil[2]{\orgdiv{Guangxi Key Lab of Multi-Source Information Mining and Security}, \orgname{Guangxi Normal University}, \orgaddress{\city{Guilin}, \postcode{541004}, \state{Guangxi Province}, \country{China}}}

\affil[3]{\orgdiv{Media Analytics and Computing Lab}, \orgname{Xiamen University}, \orgaddress{\city{Xiamen}, \postcode{361005}, \state{Fujian Province}, \country{China}}}



\abstract{The success of visual tracking has been largely driven by datasets with manual box annotations. However, these box annotations require tremendous human effort, limiting the scale and diversity of existing tracking datasets. In this work, we present a novel high-performance Self-Supervised Tracking model named \textbf{{\tracker}}, designed to eliminate the need of box annotations. Specifically, we design an effective weak-to-strong self-supervised training framework that aims to narrow the feature distribution gap between labeled and unlabeled frames. Building upon this framework, a decoupled spatio-temporal consistency training strategy is introduced to capture rich target information across timestamps by leveraging global spatial localization and local temporal association. Then, a self-prompting evolution module is designed to jointly mine the target appearance evolution at both the feature and decision levels, enabling more robust adaptation to complex and diverse unlabeled tracking scenarios. Furthermore, an instance contrastive loss is formulated to learn instance-level correspondences from a multi-view perspective, providing robust instance-level supervision without any additional annotations. This new design paradigm enables {\tracker} to not only learn a generalizable tracking representation in a low-annotation self-supervised manner, but also makes it possible to simulate realistic appearance and motion variations of target instances in real-world scenarios. Extensive experiments on ten benchmark datasets demonstrate that {\tracker} surpasses \textit{SOTA} self-supervised tracking methods, achieving an improvement of more than 25.8\%, 21.4\%, and 15.8\% in AUC (AO) score on the GOT10K, LaSOT, TrackingNet datasets, respectively, thereby significantly narrowing the performance gap with fully supervised trackers. Code is available at \href{https://github.com/GXNU-ZhongLab/SSTrack}{here}.
}




\maketitle

\section{Introduction}\label{intro}

   \begin{figure}[t]
      \centering
      \includegraphics[width=1\columnwidth]{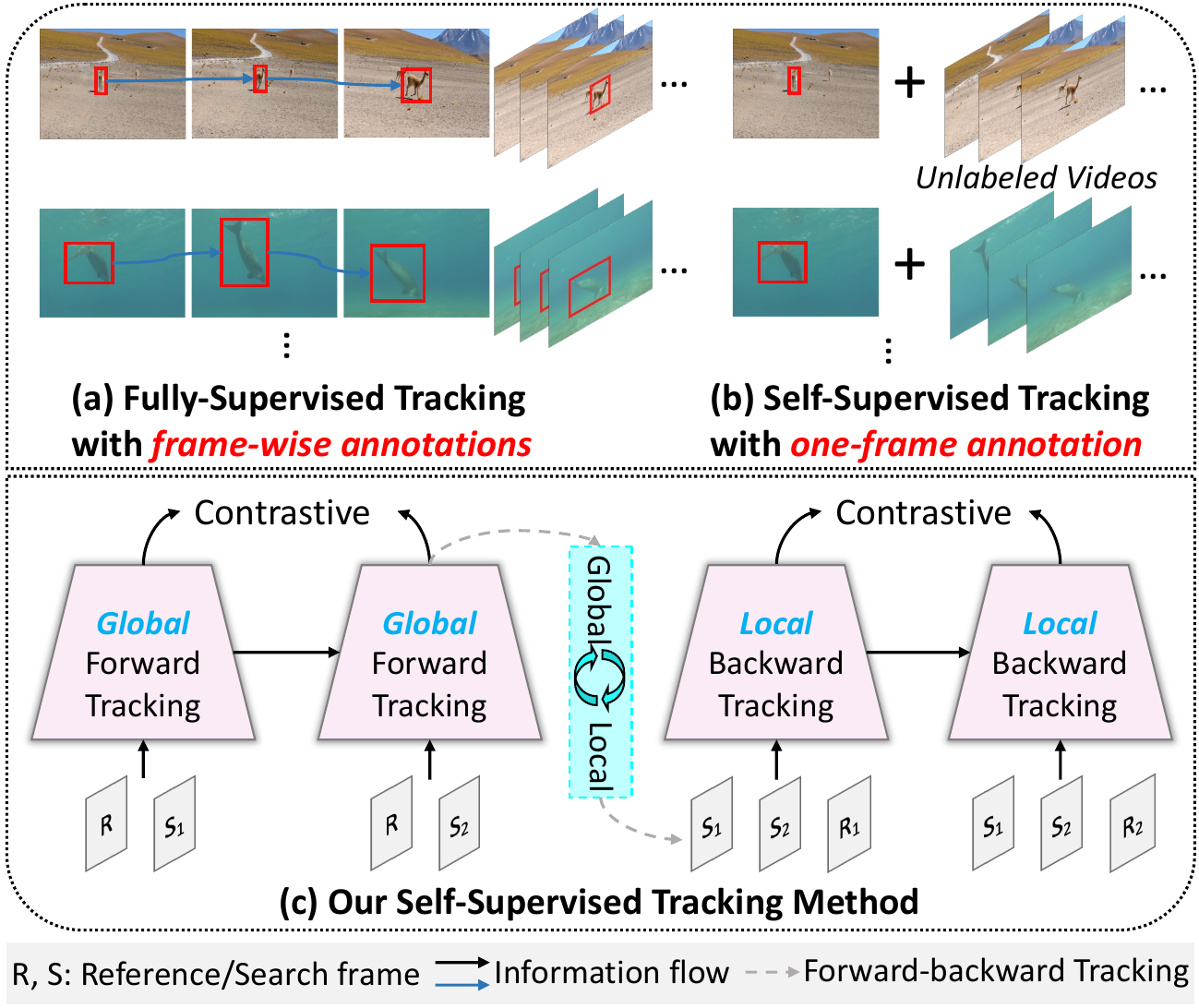}
       \caption{The annotation requirements of different tracking tasks and our proposed framework. (a) The fully-supervised tracking methods \cite{SiamRPN++,transt} with frame-wise annotations. (b) The self-supervised tracking methods \cite{s2siamfc,SDCT} with one-frame annotation. (c) Our self-supervised tracking method based on decoupled spatio-temporal consistency learning strategy and instance contrastive loss.}
       \label{fig:motivation}
    \end{figure}

Given an arbitrarily initial target, visual object tracking (VOT) requires recognizing and tracking an object in subsequent video frames. With the rapid advancement of visual object tracking technology, it has been widely applied in areas such as video surveillance, video understanding, autonomous navigation, and robotic vision.

To achieve the research goal of high-performance tracking, current visual tracking algorithms are typically trained using the full bounding box annotations of published tracking datasets \cite{lasot,trackingnet,coco,got10k}, as shown in Fig.\ref{fig:motivation}(a). 
However, as the field advances toward autonomous tracking, existing visual tracking paradigms face two fundamental limitations: (i) the lack of low-cost self-supervised frameworks, with most trackers still relying heavily on large-scale annotated data, and (ii) the inability of current self-supervised tracking methods to effectively capture accurate spatio-temporal evolutionary relationships from unlabeled videos.

   \begin{figure}[t]
      \centering
      \includegraphics[width=1\columnwidth]{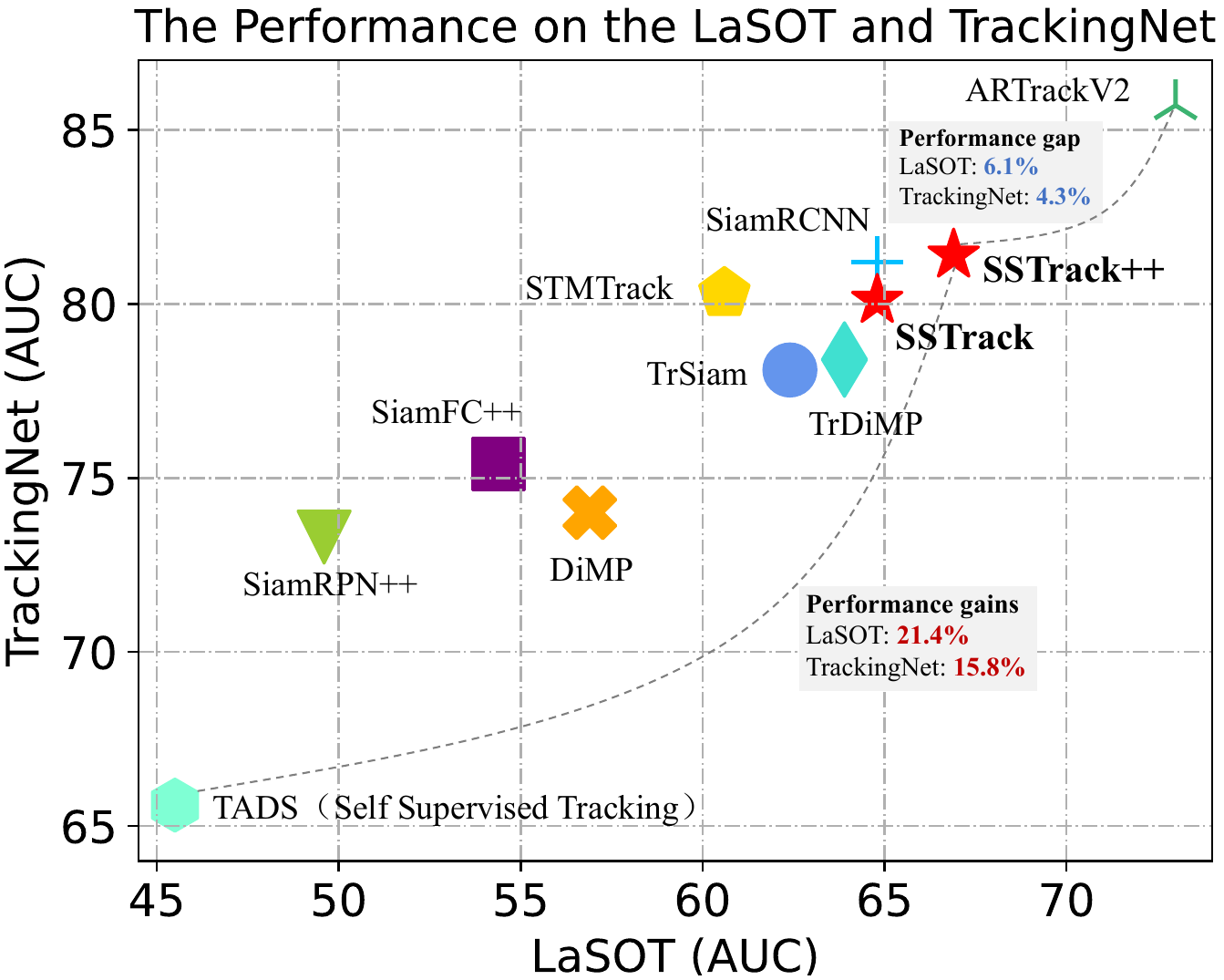}
       \caption{Comparing our {\tracker} with other tracker on the LaSOT and TrackingNet.}
       \label{fig:performance}
    \end{figure}

In practice, video data can be easily and economically collected. However, due to the prohibitive cost of manual annotation, only a few videos are annotated with bounding boxes. This constraint severely limits the scalability and diversity of existing VOT benchmarks. This further poses a significant challenge for transformer-based fully supervised tracking algorithms \cite{ostrack,ARTrack,odtrack}, which typically require massive amounts of training data to achieve optimal performance.

On the other hand, to further improve self-supervised tracking performance, a natural way is to transfer the query-based temporal association paradigm \cite{ARTrack,AQATrack,UMODTrack} to the self-supervised tracking framework. However, we find that the conventional query-based paradigm exhibits inherent limitations in this setting (see Sec.\ref{study}). Its core weakness lies in the absence of a feature evolution (purification) mechanism. Simple content-agnostic query vectors are incapable of extracting reliable spatio-temporal representations from unlabeled videos overwhelmed by noise, particularly in long-term and complex tracking scenarios.

From these two perspectives, enabling a model to \textit{automatically learn} and \textit{associate} tracked instances directly from unlabeled videos becomes essential for advancing visual tracking.
Therefore, as shown in Fig.\ref{fig:motivation}(b), we reconsider the need for box annotations by exploring \textit{a new spatio-temporal evolutionary self-supervised tracking method under an initial bounding box setting}.

To minimize the reliance on box annotations, some self-supervised tracking methods \cite{cycle_consistency,s2siamfc,SDCT,TADS} have been proposed to learn object correspondences from unlabeled videos. They learn instance tracking representations through contrastive learning or cycle-consistency matching strategies.
For instance, $S^2$SiamFC \cite{s2siamfc} and TADS \cite{TADS}, based on supervised tracking methods, generate training pairs through random region selection and data augmentation to train self-supervised trackers. Meanwhile, self-SDCT \cite{SDCT} and CycleSiam \cite{cycleSiam} rely on the principle of cyclic consistency to construct a self-supervised tracking framework with forward-backward alignment.
Despite these studies performing well in most tracking scenarios, they still face a significant performance bottleneck \textit{due to the difficulty in effectively leveraging the rich spatio-temporal context and instance correspondence in continuous video frames}.

In this work, we propose a novel self-supervised visual tracking framework, called \textbf{\tracker}, which aims to eliminate the need for expensive manual annotations while efficiently injecting spatio-temporal contextual information.
As shown in Fig.\ref{fig:motivation}(c), we reconsider the design of the self-supervised tracking framework from a new perspective. Unlike fully supervised methods \cite{HIPTrack,odtrack} that capture context through multi-frame inputs, directly learning temporal context in self-supervised tracking is a significant challenge due to the lack of annotated video data.

To address this challenge, we design an efficient weak-to-strong self-supervised training framework that aims to narrow the feature distribution gap between labeled and unlabeled frames, thereby fully exploiting the abundant unlabeled video data. Building upon this framework, a decoupled spatio-temporal consistency training strategy is introduced to automatically learn rich target information across timestamps. Specifically, we first perform forward tracking, globally searching for the spatial position of object. Then, we conduct backward tracking, locally perceiving the appearance and motion changes of the instance. Through this decoupled learning approach, we achieve global spatial localization and local temporal association within a unified framework, thereby effectively utilizing both labeled and unlabeled video data.

Furthermore, since the appearance and motion patterns of a target evolve continuously over time, we argue that target reference prompts in long-term tracking scenarios should also be dynamically updated.
Thus, a self-prompting evolution module is introduced to jointly perceive target instance information within unlabeled video frames at both the feature and decision levels, enabling deeper exploration of fine-grained appearance variations and improving adaptability to complex and diverse unlabeled tracking scenarios.

Furthermore, we introduce an instance contrastive loss function to learn instance-level correspondence across views, providing robust instance supervision without any labels.
This new design paradigm enables {\tracker} to not only learn a generalizable tracking representation in a low-annotation self-supervised manner, but also makes it possible to simulate realistic appearance and motion variations of target instances in real-world scenarios.
As shown in Fig.\ref{fig:performance}, compared to our conference version, {\tracker} achieves a 2.1\% improvement in AUC score on the LaSOT.

In summary, {\tracker} is a high-performance self-supervised tracking framework. Extensive experiments on ten tracking benchmarks (including LaSOT \cite{lasot}, TrackingNet \cite{trackingnet}, GOT10K \cite{got10k}, LaSOT$_{\rm{ext}}$ \cite{lasot-ext}, VOT2020 \cite{VOT2020}, VOT2018 \cite{VOT2018}, TNL2K \cite{tnl2k}, UAV \cite{uav123}, VastTrack \cite{vasttrack} and OTB100 \cite{OTB2015}) show that our method achieves excellent tracking performance with limited annotations and significantly narrows the performance gap between self- and fully supervised tracking methods. The main contributions are as follows:

\begin{itemize}
    \item To the best of our knowledge, we present the first self-supervised tracking model that incorporates target-aware spatio-temporal evolution, aiming to explore autonomous tracking capability under weak supervisory signals.

    \item Building on the weak-to-strong training framework, we propose a self-prompting evolution module that effectively mines accurate target appearance and motion information from unlabeled videos, enabling robust instance association in complex tracking scenarios.

    \item We propose a decoupled spatio-temporal consistency learning strategy that learns representation invariance through global spatial localization and local temporal association. Meanwhile, an instance contrastive loss is introduced to learn instance correspondences from a multi-view perspective without requiring any labels.
    
\end{itemize}

This paper builds upon our conference paper \cite{SSTrack} and significantly extends it in various aspects.
First, we provide more in-depth discussions regarding motivations and implementations of our dynamic spatio-temporal association self-supervised tracking method.
Second, built upon the weak-to-strong training framework and the self-prompting evolution module, {\tracker} is designed to accurately perceive fine-grained appearance variations across frames, further enhancing self-supervised tracking performance.
Third, more experiments and ablation studies are designed to demonstrate the effectiveness of the core components of our model, especially focusing on the proposed self-prompting evolution module.
Finally, more visual results are added for providing a comprehensive and intuitive understanding of the performance of our model in complex unlabeled tracking scenarios.

\section{Related Work}\label{work}

\subsection{Fully-Supervised Tracking}
The prevailing visual tracking algorithms \cite{SiamFC,Siamban,stark,ostrack,SATrack,DyHiT,zhao2024spatio} predominantly adhere to the fully supervised tracking paradigm and achieve high performance by training on large-scale labeled datasets such as LaSOT \cite{lasot}, TrackingNet \cite{trackingnet}, COCO \cite{coco}, and GOT10K \cite{got10k}.
These supervised tracking algorithms can be broadly categorized into two types: Siamese tracking framework \cite{SiamFC,SiamRPN,Siamban,SiamPIN} and Transformer tracking framework \cite{ostrack,mixformer,MMTrack,ARTrack,odtrack}. The former typically follows a three-stage approach involving feature extraction, fusion, and bounding box prediction for visual tracking, while the latter generally employs a transformer network to simultaneously perform feature extraction and fusion.
Finally, a prediction head network is used to perform real-time localization of the target.

Benefiting from training datasets with thousands of manual bounding box annotations, these methods have achieved significant performance gains. However, constructing large-scale video datasets is exceedingly time-consuming and costly, making it challenging to keep pace with rapid advancements in supervised tracking algorithms and often leading to gaps between data distributions in real-world scenarios. Essentially, the availability of large-scale, high-quality datasets is increasingly becoming a bottleneck for the progress of supervised tracking. Thus, there is an urgent need to research a novel and effective self-supervised tracking to alleviate this problem.

\subsection{Self-Supervised Tracking}
Unlike mainstream fully-supervised tracking algorithms, self-supervised tracking algorithms \cite{s2siamfc,SDCT,TADS} face greater challenges due to the lack of sufficient supervision signals. Current self-supervised tracking frameworks are typically divided into those based on cycle consistency and those based on contrastive learning methods.
1) \textit{Cyclic-consistency based self-supervised tracking methods}. 
self-SDCT \cite{SDCT} introduces a multi-cycle consistency loss and low similarity dropout strategy to train the feature extraction network, enhancing the robustness of the self-supervised tracker.
CycleSiam \cite{cycleSiam} leverages cycle-consistent techniques, along with region proposal and mask regression networks, to explore a Siamese self-supervised tracking framework that simultaneously performs tracking and segmentation tasks.
2) \textit{Contrastive learning based self-supervised tracking methods}. 
$S^2$SiamFC \cite{s2siamfc} randomly selects a region of the image and a corresponding enlarged region as a sampling pair, and proposes adversarial appearance masking technique for self-supervised tracking. However, this training strategy tends to sample low-quality sample pairs and fails to utilize temporal information from multiple consecutive frames.
TADS \cite{TADS} proposes a generalized data augmentation technique such as crop-transform-paste operation and is based on several supervised tracking frameworks \cite{SiamRPN++,transt} to train high performance self-supervised trackers.

Inspired by the above studies, researching visual tracking algorithms with minimal supervision signals emerges as a highly promising direction. However, unlike these works, we introduce a new self-supervised training framework from the perspective of decoupled spatio-temporal modeling, which avoids the impact of low-quality sample pairs on the training process. Furthermore, we propose a novel baseline method named {\tracker}, focusing on unlocking the potential of self-supervised tracking by collecting and correlating real-time target information across timestamps and views.

\subsection{Efficient Token Pruning in LLMs}

Efficient Token Pruning has become one of the pivotal techniques in large language models (LLMs), aiming to improve inference efficiency while maintaining stable performance. A series of studies have investigated token compression and merging strategies. For example, DynamicViT \cite{DynamicViT} and EViT \cite{EViT} dynamically select a subset of informative tokens from the original image tokens to reduce computational cost, whereas ToMe \cite{ToMe} accelerates inference by merging similar tokens and eliminating redundancy. These approaches mainly focus on token pruning in the spatial domain and have demonstrated effectiveness in various multi-modal reasoning tasks. In contrast, our work does not revisit low-score token sparsification \cite{ostrack} for visual object tracking. Instead, we focus on designing an efficient self-prompting temporal evolution mechanism that extracts high-quality target tokens to facilitate cross-frame association in unlabeled tracking scenarios.

\section{Methodology}\label{method}

\subsection{Overview}

We extend the previous SSTrack model \cite{SSTrack} to {\tracker}, a simple yet effective transformer-based self-supervised tracking framework. By optimizing the training architecture and exploiting spatio-temporal memory more efficiently, {\tracker} achieves improved self-supervised tracking performance while maintaining the same computational efficiency. In this section, we first revisit the definition of the self-supervised visual tracking task and elaborate on the three key components of {\tracker}: the decoupled spatio-temporal consistency learning strategy, the self-prompting evolution module, and the instance contrastive loss. We then introduce a weak-to-strong self-supervised training framework.

\subsection{Task Definition}

Given an unlabeled RGB video of T frames, $\mathcal{V}=\{ I_{1}, I_{2}, ..., I_{t} \} \in \mathbb{R}^{T \times H \times W \times 3} $, and an initial frame $I_{0}$ annotated with a bounding box, the goal of self-supervised tracking is to \textit{train a tracker from completely unlabeled videos} and accurately localize the target instance in subsequent frames. To this end, we introduce a novel transformer-based weak-to-strong self-supervised tracking model, $\mathcal{E}(\cdot;\theta)$, that localize the instance frame-by-frame with a dynamic spatio-temporal memory $\mathcal{M}$ used to propagate temporal cues along the video sequence.

\subsection{Tracking Background}

In order to comprehensively understand our novel self-supervised tracking framework, it is necessary to summarize the previous mainstream fully supervised tracking methods.

Despite differences in technical solutions, nearly all top-performing fully supervised methods are based on a common principle: embedding paired frame and bounding box, such as $(I_r, B_r)$, into the tracking network $\mathcal{E}$. As a result, we summarize the fully supervised tracking as follows:
\begin{equation}
B_s = \mathcal{E}\left(I_s, \left\{(I_r, B_r)\right\}_n \right),
\label{eq:fully_supervised}
\end{equation}
where $B_s$ is the bounding box predicted for a given search frame $I_s$. $\left\{(I_r, B_r)\right\}_n$ represents a pair (initial) or multiple pairs of frames and bounding boxes. Then, $\left\{(I_r, B_r)\right\}_n$ are used to guide the localization of the search frame $I_s$.

Although the field of visual tracking is dominated by fully-supervised algorithms, exploring visual tracking algorithms based on other or minimal supervision signals is a highly promising research direction, as it offers the potential to eliminate dependency on labeled data. Therefore, we introduce {\tracker}, a new self-supervised tracking method based on a weak-to-strong training framework. Furthermore, we design a decoupled spatio-temporal consistency learning strategy, a self-prompting evolution module, and an instance-contrast loss function to achieve high-performance self-supervised tracking.

   \begin{figure*}
      \centering
      \includegraphics[width=1\linewidth]{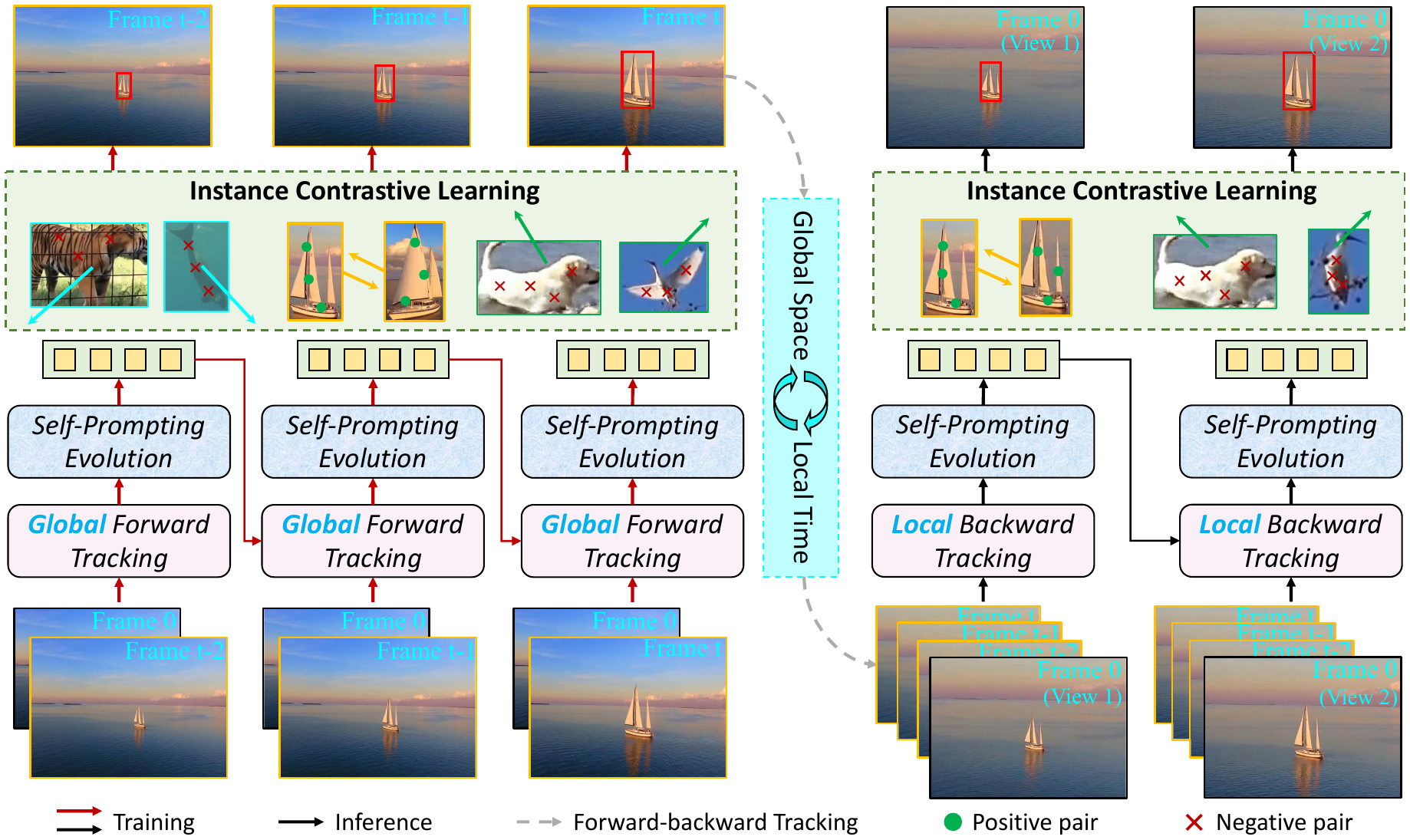}
       \caption{{\tracker} pipeline. 
       1) Forward Tracking: Given an initial frame and a global search frame, our method performs a global search to identify potential target locations.
       2) Backward Tracking: We apply local cropping and data augmentation to the original image pairs, generating two video clips with different views as inputs to our model. 
       3) Self-Prompting Evolution (SPE): SPE evaluates the quality of target prompts through model outputs and utilizes the model's own understanding of the tracking environment to optimize prompts.
       4) Instance Contrastive Learning: An instance contrastive loss is introduced to learn the similarity between different instances, achieving a robust instance tracking representation.
       }
       \label{fig:framework}
    \end{figure*}

\subsection{Decoupled spatio-temporal Consistency Learning Strategy}

Accurately obtaining target identity information is crucial for a self-supervised tracking framework. A straightforward approach is to randomly crop diverse local regions as training sample pairs, aligning with the configuration of traditional local trackers \cite{s2siamfc}. However, this method tends to produce low-quality sample pairs and fails to effectively leverage the rich spatio-temporal context, thus becoming a performance bottleneck for self-supervised tracking algorithms.
In contrast, we believe that a good self-supervised tracking algorithm should automatically locate the target instance from a global region, without being limited to a local tracking setup. As a result, as shown in Fig.\ref{fig:framework}, we propose a novel decoupled spatio-temporal consistency learning strategy that seamlessly switches between global and local tracking, automatically identifying and locating instances.

To simplify the model design, we do not add an additional global tracker but instead use a shared ViT \cite{vit} as the fundamental tracking network. Specifically, we decouple forward-backward tracking into global and local tracking.
In the forward tracking stage, given an initial frame $I_r \in \mathbb{R}^{3 \times H_r \times W_r}$ and an uncropped/full search frame $I_s \in \mathbb{R}^{3 \times H_s \times W_s}$, the ViT receives them and performs joint feature extraction and fusion to globally search for the potential target's spatial location.
Based on the current tracking results, we then crop the full search frame online to match the size of the template frame, which serves as the template frame for backward tracking. Simultaneously, we apply data augmentation operations, such as scaling, shearing, and blurring, to the initial frame to generate multiple video frames with different views, which serve as the search frames for backward tracking. This approach allows our model to simulate and learn the diverse appearance changes of target instance in real-world scenarios during backward tracking, achieving temporal cross-frame association.

With this decoupled learning way, we achieve global spatial localization and local temporal association within a unified tracking framework. This effectively leverages both labeled and unlabeled video data, making it easy to learn target correspondences across timestamps from diverse scenarios.
It is worth noting that the proposed decoupled spatio-temporal consistency learning strategy is used only during the training phase. To improve inference efficiency, we retain only the local (backward) tracking component for the inference model.

\subsection{Self-Prompting Evolution Module} \label{SPE}

In long-term and highly dynamic environments, enabling self-supervised models to effectively understand and execute complex tracking task remains a significant challenge.
Since the appearance and motion patterns of a target evolve continuously over time, we argue that target reference prompts in long-term tracking scenarios should also be dynamically updated.
To this end, we propose a novel and efficient self-prompt evolution (SPE) module. The core idea of SPE is to evaluate the quality of target appearance prompts by leveraging both feature-level and decision-level outputs produced by the self-supervised model for each video frame, and to optimize these prompts based on the model’s own understanding of the tracking environment, thereby enabling more robust adaptation to complex unlabeled tracking scenarios.

After {\tracker} performs inference on each video frame and produces the corresponding predictions, we employ the self-prompt evolution module to adaptively evaluate and select high-quality target reference prompts, progressively evolving their representations along the temporal dimension. Specifically, the prompts consist of two types: 1) fine-grained local target prompts, and 2) global tokens that integrate information across multiple timestamps. The evolution process is carried out in the following two steps:

\textbf{Step 1: Target Self-Prompt Generation.}
Although self-supervised tracking models can produce relatively accurate predictions for each video frame, the rich tracking information contained in these outputs is often underutilized. In this step, we extract and generate appearance prompts for the target instance by jointly exploiting feature-level and decision-level outputs. Specifically, to accurately capture fine-grained appearance cues of the target object, we first perform cross-attention between the center-point feature of the template frame and the search features, yielding an importance weight distribution over the current search tokens. Since this importance distribution reflects the spatial likelihood of the target’s presence, we interpret high similarity scores as corresponding to target regions, while low similarity scores indicate background regions. Accordingly, we introduce a filtering ratio $\lambda$ to discard a small portion of background tokens in advance, thereby reducing interference. Concretely, search tokens with low similarity scores within the filtering ratio $\lambda$ are marked as invalid and recorded in a mask matrix $\mathcal{M}$. 
In other words, this mask is a binary matrix indicating whether each search token is discarded, and is used to pre-filter low-quality background tokens.
Finally, we jointly compute cross-correlations among the decision-level classification confidence map $\phi$, the importance weight distribution $\mathcal{W}$, and the mask matrix $\mathcal{M}$, which together serve as a comprehensive criterion for generating target self-prompts. The detailed generation process is described as follows:

\begin{equation}
    \begin{split}
    \mathcal{W} &= \text{Softmax} \left(  \frac{\mathcal{F}_{rc} \times \mathcal{F}_{s}^\top}{\sqrt{d}}  \right), \\
    \mathcal{P} &= \text{TopK} \left( \mathcal{F}_{s}, \mathcal{M} \times \phi \times \frac{1}{n}\sum_{j=1}^{n}{\mathcal{W}} \right), \\
    \end{split}
    \label{eq:gen_prompt}
\end{equation}
where $\mathcal{F}_{rc}$ denotes the center-point feature of the template frame. $\mathcal{F}_{s}$ refers to the search frame feature. $d$ is the feature dimension. $\mathcal{P}$ denotes the selected high-quality target prompt consisting of $k$ tokens. $n$ corresponds to the number of attention heads.
Through this joint feature-level and decision-level evaluation criterion, we are able to filter out information-rich target appearance tokens from cluttered backgrounds and use them as self-prompts to be propagated to subsequent video frames, thereby enhancing the stability of self-supervised training.

\textbf{Step 2: Target Self-Prompt Evolution.}
For each unlabeled video frame, {\tracker} collects the most recent target prompts $\mathcal{P} = [\mathcal{P}_{t=1}, \mathcal{P}_{t=2},... ,\mathcal{P}_{t=n}]$ and propagates them across frames. However, in open tracking scenarios, the presence of scale variations, distractors with similar appearances, and occlusions necessitates a cross-temporal assessment of the quality of each visual token to robustly adapt to complex scene dynamics. In other words, beyond focusing on the latest target observations, it is also essential to explicitly account for historical target prompts.

To update a clean and reliable set of target prompts, we select the top $k-1$ most informative raw target tokens from the union of tokens in the current frame and historical frames according to the aforementioned evaluation criterion, and use them as the new target prompts to strengthen guidance in unlabeled tracking scenarios. Meanwhile, instead of discarding suboptimal target tokens, we further evolve them by aggregating their attention scores into a new global target prompt, thereby enhancing its representational capacity for self-supervised tracking.


Through this novel self-supervised prompt selection and evolution strategy, we effectively preserve beneficial features while gradually optimizing those with lower information content. The formulation is given as follows:

\begin{equation}
    \begin{split}
    \mathcal{P}_{new} &= \text{Mean} \left( \mathcal{W}_{n-k+1:} \times \mathcal{P}_{n-k+1:} \right), \\
    \end{split}
    \label{eq:evolution}
\end{equation}
where $\mathcal{P}_{new}$ represents a new evolutionary target prompt.

\subsection{Instance Contrastive Learning}

Another performance bottleneck of traditional self-supervised tracking algorithms is the difficulty in learning a robust instance representation. A mature visual tracker must accurately locate the target from diverse and complex backgrounds, requiring the ability to distinguish between different instances. However, directly applying contrastive learning methods \cite{MOCO,InfoNCE,simsiam} in self-supervised tracking to improve feature discriminability is not feasible, as they rely on clean, target-centered images. In other words, due to the highly open nature of instances and scenarios in the tracking domain, frame-level similarity learning is insufficient to distinguish instances in complex scenarios. Therefore, we introduce a simple yet effective instance contrastive loss function that mines rich instance information from a large number of unlabeled video sequences to efficiently learn diverse instance correspondences.

Given an initial frame $I_r \in \mathbb{R}^{3 \times H_r \times W_r}$, we apply various data augmentation operations to generate different views, simulating the appearance changes of the object at different timestamps in the video, thereby automatically obtaining target correspondences. Since annotations are unknown in the contrastive learning process, we design an additional mask matrix $\mathcal{M}$ for each view based on the predicted bounding box coordinates to extract the target instance from the background, where 1 represents the target region and 0 represents the background region. 
It is worth noting that this mask matrix differs from the one in Step 1 of Section \ref{SPE}. Specifically, the former is used to extract target instance features for instance contrastive learning, whereas the latter serves to filter out low‑quality background tokens.

Subsequently, we perform a pooling operation to obtain the corresponding target representation, achieving instance supervision without any labels. Formally, our instance-level contrastive loss function is as follows:
\begin{equation}
    \resizebox{.9\hsize}{!}{
    $\mathcal{L}_{cont} = -\sum_{q \in Q} \log \frac{\exp\left(\text{sim}(q, q^+) / \tau \right)}{\sum_{q^- \in Q^-} \exp\left(\text{sim}(q, q^-) / \tau \right)},$
}
 \label{eq:contloss}
\end{equation}
where $Q$ represents the set of all potential instances in a batch. $q^+$ and $q^-$ denote the positive and negative samples to $q$, respectively. Positive samples are different views of the same instance obtained through various data augmentation operations. Negative samples come from different instances. Additionally, $\text{sim}(\cdot)$ denotes the cosine similarity between any sample pairs and $\tau$ is a temperature parameter.

It is worth noting that when computing the loss, we apply a filtering strategy on the predicted bounding boxes: samples with invalid or extremely small box areas (e.g., below a predefined threshold of 0) are excluded from the contrastive learning process, thereby preventing degenerate representations of tracking instances.

Furthermore, the proposed instance contrastive learning module operates on feature-level similarity rather than hard ground truth box-level supervision, making the optimization substantially more tolerant to localization noise. The mask introduced in this section serves as a coarse validity constraint to remove degenerate regions and is not the sole mechanism for suppressing noisy target instance embeddings. More importantly, the proposed model is jointly constrained by multiple modeling techniques, including the decoupled spatio-temporal consistency learning strategy and the self-prompting evolution module. As joint training progresses under multiple modeling techniques, the temporal associations gradually become more reliable, leading to progressively refined instance embeddings and improved predicted bounding box quality.

Through this learning process, we make the representations of the same instance as similar as possible in the feature space while maximizing the distance between representations of different instances. This allows our model to effectively learn robust instance tracking representations in a self-supervised manner, reducing reliance on extensive box annotations.

\subsection{Weak-to-Strong Self-Supervised Training Framework}

Previous self-supervised tracking methods \cite{SDCT,cycleSiam} often adopt a training framework based on forward-backward (temporal) cycle consistency \cite{cycle_consistency}. However, these methods are constrained by the optimization of the backward tracking phase, leaving the forward tracking process unsupervised and preventing full utilization of unlabeled video.

\begin{figure}[t]
      \centering
      \includegraphics[width=1\columnwidth]{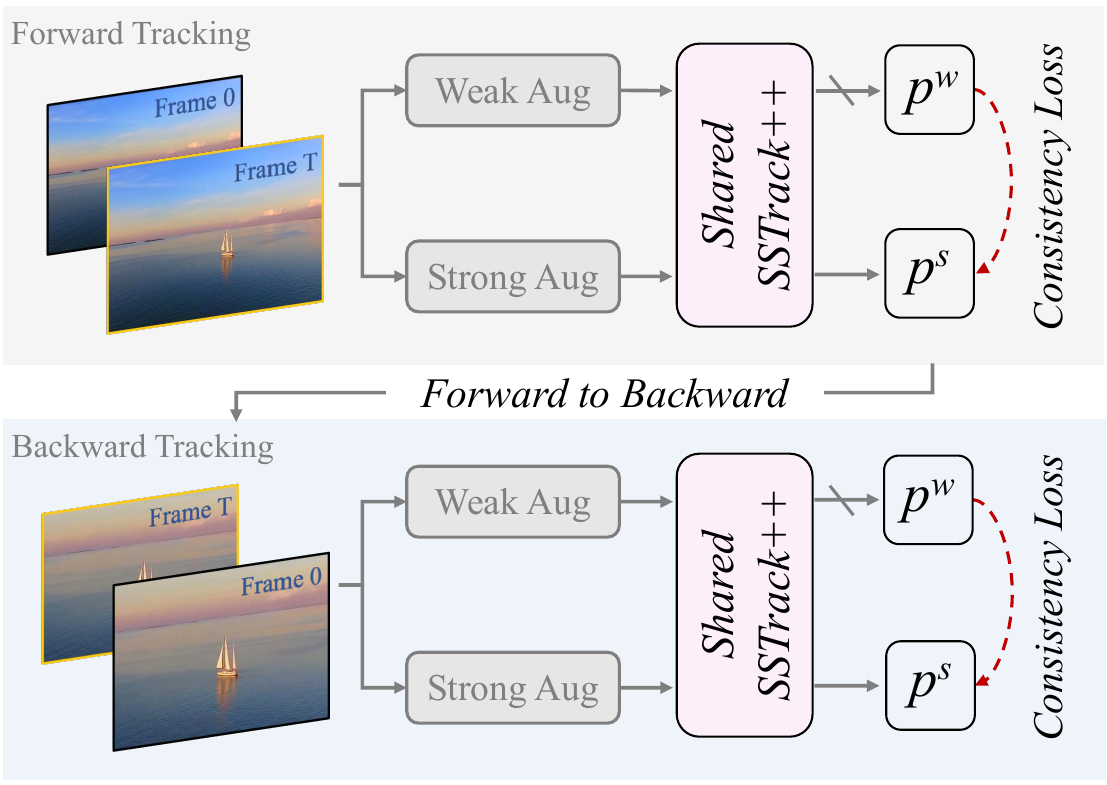}
       \caption{The weak-to-strong self-supervised training framework. The weak sample branch is set to a gradient stop state, preventing its parameters update.}
       \label{fig:concept}
    \end{figure}

Inspired by the easy-to-hard learning philosophy \cite{fixmatch}. The key insight is that predictions generated from weakly augmented inputs (with less noise) exhibit higher reliability and lower variance, and can thus serve as supervisory signals to guide target localization under strongly augmented inputs. Consequently, in the self-supervised video object tracking task, to better exploit unlabeled video data, we design a novel weak-to-strong self-supervised training framework consisting of a weak tracking branch and a strong tracking branch, as shown in Fig.\ref{fig:concept}. Specifically, the outputs of the weak branch are treated as pseudo bounding-box labels to supervise the strong branch, enabling it to learn accurate localization under more challenging conditions. 
Theoretically, our self-supervised framework jointly learns bounding box decoding and general tracking representation from unlabeled videos. It not only leverages the powerful idea of cyclic consistency training strategy, but also inherits the advantages of contrastive learning in general representation learning.

In the {\tracker} framework, we generate a strongly augmented frame at each time step through data augmentation operations (i.e., horizontal flipping and random jitter), forming a strong–weak sample pair to construct the temporal input sequence $\{I_{s}^{t}, I_{s}^{t'} \}$. The video sequence is then fed into the model in a temporal auto-regressive manner, and the learning process can be formulated as follows:
   \begin{equation}
     \begin{split}
     B_{s}^{t'}, \sigma^{t'} &= \mathcal{E} \left(I_{s}^{t'} | \left(I_s^{t}, B_{s}^{t}, \sigma^{t} \right), \left\{ \left(I_r, B_r\right) \right\}_n \right).
      \end{split}
     \label{eq:auto_regressive}
   \end{equation}
Here, $I_s^{t}$ and $B_{s}^{t}$ denote the weakly augmented frame and its predicted bounding box at time step $t$, while $I_s^{t'}$ and $B_{s}^{t'}$ correspond to the strongly augmented frame and its predicted bounding box at the same time step. $\sigma^{t'}$ and $\sigma^{t}$ denote the temporal token vectors extracted from the strongly and weakly augmented video frames, respectively.

During training, the weak sample (i.e., the original frame $I_{s}^{t}$) first predicts the target bounding box through the workflow $I_{s}^{t} \rightarrow \mathcal{E} \rightarrow B_{s}^{t}$. Subsequently, the strongly augmented frame $I_{s}^{t'}$ is processed in an independent branch, where multiple transformer layers perform feature extraction and fusion to produce a new prediction $B_{s}^{t'}$. Consequently, this procedure constitutes a forward or backward tracking training pipeline.

Specifically, to easily understand the self-supervised cycle consistency tracking process, we use $\mathcal{C}$ to denote the target context learned from the reference frames $\left\{(I_r, B_r)\right\}_n$, and then we iteratively employ tracking model $\mathcal{E}$ $i$ times in a forward manner from time step $t-i$ to $t$:
   \begin{equation}
     \begin{split}
        B^{t}_s &= \mathcal{E}^{i}(I^{t}_s, \mathcal{C}) \\ 
                &= \mathcal{E}(I^{t-1}_s, \mathcal{E}(I^{t-2}_s,..., \mathcal{E}(I^{t-i}_s, \mathcal{C}))).
      \end{split}
     \label{eq:forward}
   \end{equation}

In the backward tracking phase, the tracking network $\mathcal{E}$ is employed backwards $i$ times from time step $t$ to $t-i$:
   \begin{equation}
     \begin{split}
        B^{t-i}_s &= \mathcal{E}^{i}(I^{t-i}_s, \mathcal{C}) \\
                &= \mathcal{E}(I^{t-i+1}_s, \mathcal{E}(I^{t-i+2}_s,..., \mathcal{E}(I^{t}_s, \mathcal{C}))).
      \end{split}
     \label{eq:bardward}
   \end{equation}

Based on the above formulation, we can construct a self-supervised tracking framework. We further develop optimization objectives for the proposed model to effectively learn the target correspondence from unlabeled videos.

For the initial frame with bounding box annotations and its synthetic samples generated by data augmentations, we apply a supervised loss $\mathcal{L}_{s}$. Specifically, this supervised term is a combination of classification, regression, and contrastive learning losses between model predictions and ground truth.

For subsequent unlabeled frames, we rely on the proposed weak‑to‑strong self‑supervised training framework to generate pseudo labels, which are then used to compute the unsupervised loss $\mathcal{L}_{u}$. Specifically, this unsupervised term regularizes the prediction of the sample under strong augmentations to be the same as that under weak augmentations. Notably, gradients are stopped for the weak sample branch, preventing its parameters from being updated. Meanwhile, $\mathcal{L}_{u}$ is also a combination of classification and regression losses.
Furthermore, to narrow the feature distribution gap between labeled and unlabeled frames, we calculate the mean squared error (MSE) between strong and weak temporal tokens.

The tracking optimization objectives can be formulated as:
\begin{equation}
   \begin{split}
      \mathcal{L}_{s} &= \mathcal{L}_{cls}(B_s, B^{gt}_s) + \mathcal{L}_{reg}(B_s, B^{gt}_s) + \mathcal{L}_{cont}, \\
      \mathcal{L}_{u} &= \mathcal{L}_{cls}(B_s^{t'}, B^{t}_s) + \mathcal{L}_{reg}(B_s^{t'}, B^{t}_s),
    \end{split}
     \label{eq:trackloss}
   \end{equation}
where $\mathcal{L}_{cls}$ is the Focal loss \cite{focalloss} and $\mathcal{L}_{reg}$ denotes the combination of GIoU loss \cite{giou} and $\mathcal{L}_1$ loss.

Then, the overall objective function is a combination of supervised loss $\mathcal{L}_{s}$, unsupervised loss $\mathcal{L}_{u}$ and MSE loss $\mathcal{L}_{mse}$ as:
   \begin{equation}
     \begin{split}
     \mathcal{L}_{all} &= \omega_{1} \mathcal{L}_{s} + \omega_{2} \mathcal{L}_{u} + \omega_{3} \mathcal{L}_{mse},
      \end{split}
     \label{eq:all}
   \end{equation}

By introducing this weak-to-strong self-supervised training framework, the model can fully exploit unlabeled videos and be trained in a self-supervised manner to achieve high-performance visual tracking.

Finally, we summarize the process of the proposed self-supervised tracking algorithm as shown in the Algorithm \ref{alg:algorithm}.

\begin{algorithm}[t]
    \caption{{\tracker} training process}
    \label{alg:algorithm}
    \KwIn{Initial frame and bounding box $(I_r, B_r)$; Search frame $I^{t=2:n}_s$} 
    \KwOut{tracking model $ \mathcal{E}(\cdot) $} 
    // {\textit{\color{blue} Forward tracking}} \\
    \For{$t=2$ to $n$}  
    { 
      $B_{s}^{t'}, \sigma^{t'} = \mathcal{E} \left(I_{s}^{t'} | \left(I_s^{t}, B_{s}^{t}, \sigma^{t} \right), \left\{ \left(I_r, B_r\right) \right\}_n \right)$; \\
      Crop $I^t_s$ based on $B^t_s$ yields a new reference frame $I^t_{sr}$; \\
      Collect and evolve target self-prompts using Eq.\ref{eq:gen_prompt} and Eq.\ref{eq:evolution};
    }
    // {\textit{\color{blue} Backward tracking}} \\
    Expand $I_r$ to get multiple new views $I^{1:m}_r$; \\
    \For{$t=1$ to $m$}  
    { 
      $B_{s}^{t'}, \sigma^{t'} = \mathcal{E} \left(I_{s}^{t'} | \left(I_s^{t}, B_{s}^{t}, \sigma^{t} \right), \left\{ \left(I_r, B_r\right) \right\}_n \right)$; \\
      Collect and evolve target self-prompts using Eq.\ref{eq:gen_prompt} and Eq.\ref{eq:evolution};
    }
    // {\textit{\color{blue} Calculate losses}} \\
    Calculate loss using Eq.\ref{eq:all} and update parameters.
\end{algorithm}

\subsection{Discussion on Fully Supervised Tracking}

Due to limitations in the supervision signal, current self-supervised tracking algorithms perform relatively poorly compared to fully supervised tracking methods.
However, we argue that the research goal of self-supervised tracking should not be narrowly defined as matching or surpassing fully supervised methods. Instead, its core value lies in the following aspects:

1) Reducing annotation dependency and improving scalability. Vast amounts of unlabeled video data are readily available on the Internet. Self-supervised methods provide a viable paradigm for leveraging large-scale unlabeled data to learn spatio-temporal representations, which is beyond the scope of fully supervised approaches.

2) Enhancing cross-domain generalization for open-world scenarios. In real-world applications (e.g., Nighttime UAV tracking, underwater environments, and camouflaged scenes), data distributions often differ significantly from those of standard training datasets \cite{lasot,trackingnet}. Since self-supervised methods do not rely on specific annotation distributions, they inherently offer potential for domain generalization.

3) Furthermore, we argue that self-supervised tracking may be better suited as a foundational pre-training paradigm for visual tracking, rather than serving as a substitute for fully supervised methods.

\section{Experiments}\label{exp}
   
\subsection{Implementation Details}
   We use a ViT-Base \cite{vit} model with DropMAE \cite{DropMAE} pre-trained parameters as the visual encoder.
   The training data includes LaSOT \cite{lasot}, GOT-10k \cite{got10k}, TrackingNet \cite{trackingnet}, and COCO \cite{coco}.
   The AdamW \cite{adamw} is used to end-to-end optimize model parameters with initial learning rate of $2.5 \times 10^{-5}$ for the backbone, $2.5 \times 10^{-4}$ for the rest, and set the weight decay to $10^{-4}$.
   The training epochs is set to $150$ epochs. $10k$ image pairs are randomly sampled in each epoch.
   The learning rate drops by a factor of $10$ after $120$ epochs.
   The model is conducted on a server with two 80GB Tesla A100 GPUs and set the batch size to be $8$.
   For forward tracking, we use one reference frame and three global search frames as the model input. For backward tracking, we use three reference frames and two cropped search frames as the input. The reference and search frames for backward tracking are derived from the search and reference frames of forward tracking, with the reference frames augmented from different views.

   \begin{figure}[t]
      \centering
      \includegraphics[width=1\columnwidth]{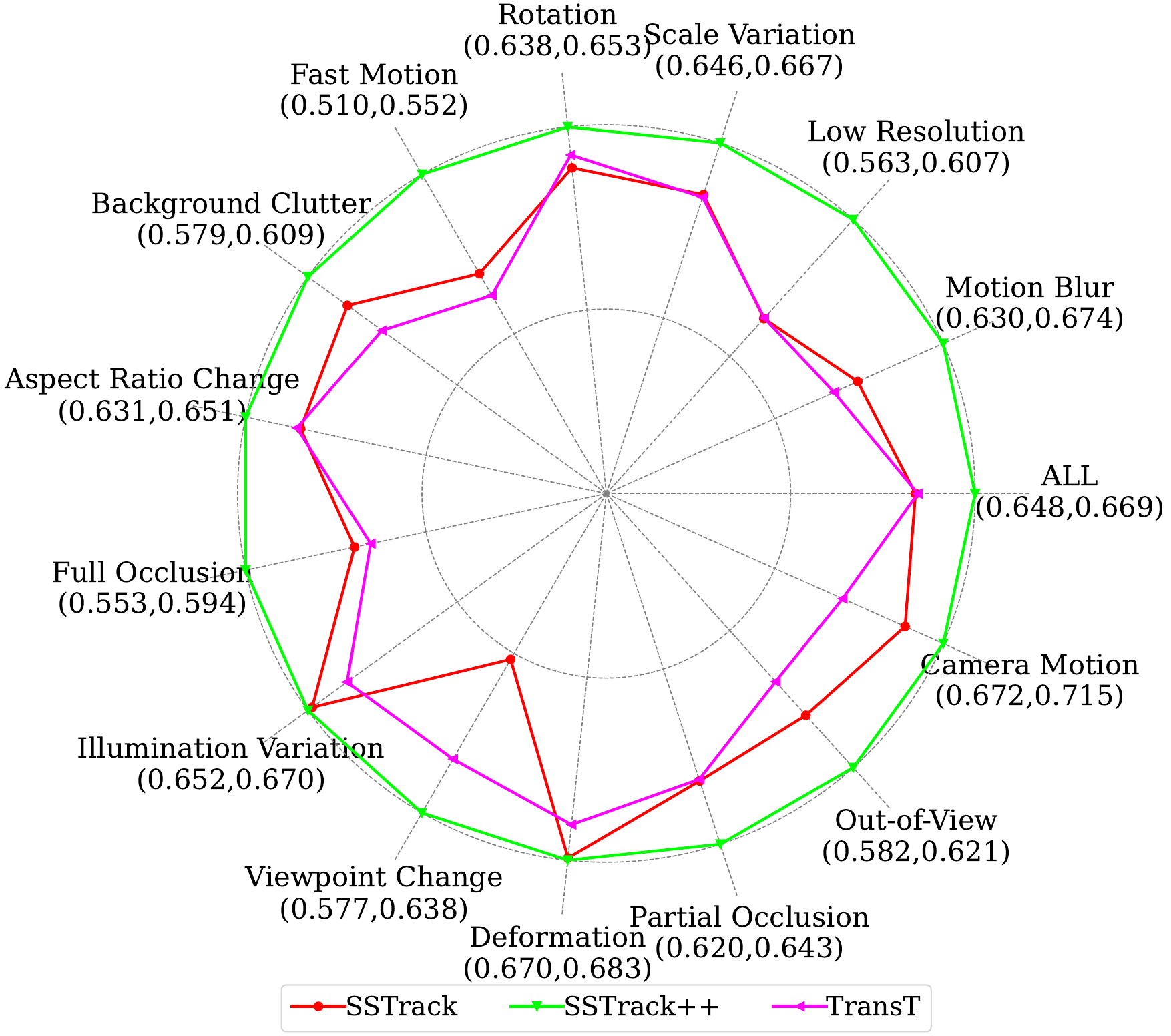}
       \caption{AUC scores of different attributes on LaSOT dataset.}
       \label{fig:attr}
    \end{figure}

\subsection{Comparison with the SOTA}
We compare the performance of our SSTrack and {\tracker} with previous self-supervised tracking methods on the GOT10K \cite{got10k}, LaSOT \cite{lasot}, TrackingNet \cite{trackingnet}, OTB100 \cite{OTB2015}, UAV123 \cite{uav123}, and VOT2018 \cite{VOT2018} datasets. We then compare our two trackers with more fully supervised methods on the LaSOT$_{\rm{ext}}$ \cite{lasot-ext}, TNL2K \cite{tnl2k}, VOT2020 \cite{VOT2020} and VastTrack \cite{vasttrack} datasets.

\textbf{GOT10K.}
GOT10K \cite{got10k} is a popular general tracking benchmark containing over $10,000$ video sequences. Under the one-shot protocol of the GOT10K dataset, we compare our SSTrack and {\tracker} with both self- and fully- supervised algorithms. Benefiting from the proposed weak-to-strong self-supervised training framework, we are able to stably train an effective self-supervised tracking model. As shown in the Tab.\ref{tab:results}, compared to the self-supervised method TADS \cite{TADS}, our tracker significantly outperforms by 25.7\%, 27.1\%, and 45.1\% in AO, SR${_{0.5}}$, and SR${_{0.75}}$ metrics, respectively. 
Additionally, our method significantly narrows the performance gap with fully supervised methods. This performance gain is primarily attributed to the proposed decoupled spatio-temporal consistency learning strategy, which effectively leverages both labeled and unlabeled video data to learn the spatio-temporal context of target instance.

\begin{table*}[t]
    \centering
    \caption{Comparison with state-of-the-arts on four popular benchmarks: GOT10K \cite{got10k}, LaSOT \cite{lasot}, TrackingNet \cite{trackingnet}, and LaSOT$_{\rm{ext}}$ \cite{lasot-ext}.
    \textit{Fully Sup} denotes fully supervised tracking methods, while \textit{Self Sup} refers to self-supervised tracking methods.
    Where $*$ denotes for trackers only trained on GOT10K.
    The best two results are shown in {\color{red}red} and {\color{blue}blue} fonts.}
    \label{tab:results}
    \resizebox{\textwidth}{!}{
    \begin{tabular}{c|l|ccc|ccc|ccc|ccc}
    \toprule
     \multicolumn{1}{c|}{\multirow{2}{*}{Type}} & \multicolumn{1}{c|}{\multirow{2}{*}{Method}} & \multicolumn{3}{c|}{GOT10K$^*$} & \multicolumn{3}{c|}{LaSOT} & \multicolumn{3}{c|}{TrackingNet} & \multicolumn{3}{c}{LaSOT$_{\rm{ext}}$} \\
      & & AO & SR${_{0.5}}$ & SR${_{0.75}}$ & AUC & P${_{\rm{Norm}}}$ & P & AUC & P${_{\rm{Norm}}}$ & P & AUC & P${_{\rm{Norm}}}$ & P \\
      \midrule
      \multicolumn{1}{c|}{\multirow{22}{*}{Fully Sup}} & SiamPRN++ \cite{SiamRPN++} & 51.7 & 61.6 & 32.5 & 49.6 & 56.9 & 49.1 & 73.3 & 80.0 & 69.4 & 34.0 & 41.6 & 39.6 \\
      & DiMP \cite{DiMP50} & 61.1 & 71.7 & 49.2 & 56.9 & 65.0 & 56.7 & 74.0 & 80.1 & 68.7 & 39.2 & 47.6 & 45.1 \\
      & SiamRCNN \cite{siamrcnn} & 64.9 & 72.8 & 59.7 & 64.8 & 72.2 & - & 81.2 & 85.4 & 80.0 & - & - & - \\
      & Ocean \cite{Ocean} & 61.1 & 72.1 & 47.3 & 56.0 & 65.1 & 56.6 & - & - & - & - & - & - \\
      & STMTrack \cite{STMTrack} & 64.2 & 73.7 & 57.5 & 60.6 & 69.3 & 63.3 & 80.3 & 85.1 & 76.7 & - & - & - \\
      & TrDiMP \cite{trdimp} & 67.1 & 77.7 & 58.3 & 63.9 & - & 61.4 & 78.4 & 83.3 & 73.1 & - & - & - \\
      & TransT \cite{transt} & 67.1 & 76.8 & 60.9 & 64.9 & 73.8 & 69.0 & 81.4 & 86.7 & 80.3 & - & - & - \\
      & Stark \cite{stark} & 68.8 & 78.1 & 64.1 & 67.1 & 77.0 & - & 82.0 & 86.9 & - & - & - & - \\
      & KeepTrack \cite{keeptrack} & - & - & - & 67.1 & 77.2 & 70.2 & - & - & - & 48.2 & - & - \\
      & SBT-B \cite{SBT} & 69.9 & 80.4 & 63.6 & 65.9 & - & 70.0 & - & - & - & - & - & - \\
      & Mixformer \cite{mixformer} & 70.7 & 80.0 & 67.8 & 69.2 & 78.7 & 74.7 & 83.1 & 88.1 & 81.6 & - & - & - \\
      & TransInMo \cite{TransInMo} & - & - & - & 65.7 & 76.0 & 70.7 & 81.7 & - & - & - & - & - \\
      & OSTrack \cite{ostrack} & 73.7 & 83.2 & 70.8 & 71.1 & 81.1 & 77.6 & 83.9 & 88.5 & 83.2 & 50.5 & 61.3 & 57.6 \\
      & AiATrack \cite{aiatrack} & 69.6 & 80.0 & 63.2 & 69.0 & 79.4 & 73.8 & 82.7 & 87.8 & 80.4 & 47.7 & 55.6 & 55.4 \\
      & SeqTrack \cite{seqtrack} & 74.5 & 84.3 & 71.4 & 71.5 & 81.1 & 77.8 & 83.9 & 88.8 & 83.6 & 50.5 & 61.6 & 57.5 \\
      & GRM \cite{GRM} & 73.4 & 82.9 & 70.4 & 69.9 & 79.3 & 75.8 & 84.0 & 88.7 & 83.3 & - & - & - \\
      & DropTrack \cite{DropMAE} & 75.9 & 86.8 & 72.0 & 71.8 & 81.8 & 78.1 & 84.1 & 88.9 & - & 52.7 & 63.9 & 60.2 \\
      & ARTrack \cite{ARTrack} & 75.5 & 84.3 & 74.3 & 72.6 & 81.7 & 79.1 & 85.1 & 89.1 & 84.8 & 51.9 & 62.0 & 58.5 \\
      & ODTrack \cite{odtrack} & 77.0 & 87.9 & 75.1 & 73.2 & 83.2 & 80.6 & 85.1 & 90.1 & 84.9 & 52.4 & 63.9 & 60.1 \\
      & HIPTrack \cite{HIPTrack} & 77.4 & 88.0 & 74.5 & 72.7 & 82.9 & 79.5 & 84.5 & 89.1 & 83.8 & - & - & - \\
      & AQATrack \cite{AQATrack} & 76.0 & 85.2 & 74.9 & 72.7 & 82.9 & 80.2 & 84.8 & 89.3 & 84.3 & 52.7 & 64.2 & 60.8 \\
      & ARTrackV2 \cite{Artrackv2} & 77.5 & 86.0 & 75.5 & 73.0 & 82.0 & 79.6 & 85.7 & 89.8 & 85.5 & 52.9 & 63.4 & 59.1 \\
      \midrule
      \multicolumn{1}{c|}{\multirow{5}{*}{Self Sup}} & TADS \cite{TADS} & {46.7} & {56.5} & {21.1} & {45.5} & {54.2} & {44.8} & {65.6} & {73.4} & {60.6} & - & - & - \\
      & SSTrack-256 \cite{SSTrack} & 67.1 & {76.6} & {59.1} & {64.8} & {75.2} & {69.7} & {80.1} & {86.7} & {78.9} & {46.2} & {57.8} & {52.1} \\
      & SSTrack-384 \cite{SSTrack} & \color{blue}{72.4} & \color{red}{83.6} & \color{blue}{66.2} & {65.9} & {76.4} & {70.7} & {80.4} & {86.3} & {77.9} & \color{blue}{48.5} & \color{blue}{60.9} & \color{blue}{54.5} \\
      & \textbf{{\tracker}-256} & {69.6} & \color{blue}{80.4} & {62.8} & \color{blue}{66.3} & \color{blue}{76.6} & \color{blue}{70.8} & \color{red}{81.4} & \color{red}{87.1} & \color{red}{79.1} & {46.0} & {57.5} & {51.5} \\
      & \textbf{{\tracker}-384} & \color{red}{72.5} & \color{red}{83.6} & \color{red}{66.5} & \color{red}{66.9} & \color{red}{77.8} & \color{red}{71.6} & \color{blue}{80.7} & \color{blue}{86.8} & \color{blue}{78.7} & \color{red}{49.1} & \color{red}{61.3} & \color{red}{55.8} \\
    \bottomrule
    \end{tabular} }
\end{table*}

\textbf{LaSOT.}
LaSOT \cite{lasot} is a classic long-term tracking benchmark, comprising 1120 training sequences and 280 test sequences. As shown in the Tab.\ref{tab:results}, it can be seen that our {\tracker} achieves state-of-the-arts self-supervised tracking results with an interesting self-prompting evolution module. First, compared to the self-supervised method TADS, our SSTrack improves the success, normalized precision, and precision score by 20.4\%, 22.2\%, and 25.9\%, respectively. Additionally, compared to the state-of-the-art supervised method ODTrack \cite{odtrack}, the AUC score gap of our self-supervised tracker is reduced to 7.3\%. By continuously selecting and evolving target self-prompts in long-term tracking scenarios, our {\tracker} further narrows the performance gap with fully-supervised methods. For instance, {\tracker}-384 achieves AUC and precision scores of 66.9\% and 71.6\%, respectively. These results indicate that the proposed instance contrastive loss function helps the model learn the appearance and motion information of target, significantly enhancing self-supervised tracking performance in long-term scenarios.

Furthermore, as shown in Fig.\ref{fig:attr}, we analyze the attribute results of the proposed {\tracker} across all challenges on the LaSOT dataset, including rotation, fast motion, scale variation, occlusion, and so on. Our {\tracker} consistently outperforms both the fully supervised TransT \cite{transt} and the self-supervised SSTrack tracker on multiple challenging attributes, demonstrating the generalizability of the proposed method.

\begin{table*}[t]
\centering
\caption{Comparison with sota methods on TNL2K \cite{tnl2k} benchmark. The best two results are shown in {\color{red}red} and {\color{blue}blue} fonts.
}
\label{tab:tnl2k}
\resizebox{\textwidth}{!}{
\begin{tabular}{l|cccccccccc|cccc}
\toprule
\multicolumn{1}{c|}{\multirow{3}{*}{Metrics}} & \multicolumn{10}{c|}{Fully Supervised Tracking} & \multicolumn{4}{c}{Self Supervised Tracking} \\
& \tabincell{c}{SiamFC \\ \cite{SiamFC}} & \tabincell{c}{MDNet \\ \cite{MDNet}} & \tabincell{c}{SiamRPN++ \\ \cite{SiamRPN++}} & \tabincell{c}{Ocean \\ \cite{Ocean}} & \tabincell{c}{TransT \\ \cite{transt}} & \tabincell{c}{OSTrack \\ \cite{ostrack}} & \tabincell{c}{SeqTrack \\ \cite{seqtrack}} & \tabincell{c}{ARTrack \\ \cite{ARTrack}} & \tabincell{c}{ODTrack \\ \cite{odtrack}} & \tabincell{c}{AQATrack \\ \cite{AQATrack}} & \tabincell{c}{SSTrack-256 \\ \cite{SSTrack}} & \tabincell{c}{SSTrack-384 \\ \cite{SSTrack}} & \textbf{{\tracker}-256} & \textbf{{\tracker}-384} \\
\midrule
AUC(\%) & 29.5 & 38.0 & 41.3 & 38.4 & 50.7 & 55.9 & 56.4 & 59.8 & 60.9 & 59.3 & {52.1} & \color{blue}{53.8} & {53.1} & \color{red}{54.1} \\
P(\%) & 28.6 & 37.1 & 41.2 & 37.7 & 51.7 & - & - & - & 64.5 & 62.3 & {53.3} & \color{blue}{55.3} & {54.9} & \color{red}{56.5} \\
\bottomrule
\end{tabular} }
\end{table*}

\begin{table*}[t]
\centering
\caption{Comparison with sota methods on VOT2020 \cite{VOT2020} benchmark. The best two results are shown in {\color{red}red} and {\color{blue}blue} fonts.}
\label{tab:vot2020}
\resizebox{\textwidth}{!}{
\begin{tabular}{l|cccccccccc|cccc}
\toprule
\multicolumn{1}{c|}{\multirow{3}{*}{Metrics}} & \multicolumn{10}{c|}{Fully Supervised Tracking} & \multicolumn{4}{c}{Self Supervised Tracking} \\
& \tabincell{c}{SiamMask \\ \cite{SiamMask}} & \tabincell{c}{Ocean \\ \cite{Ocean}} & \tabincell{c}{D3S \\ \cite{D3s}} & \tabincell{c}{AlphaRef \\ \cite{Alpha-Refine}} & \tabincell{c}{Ocean+ \\ \cite{ocean+}} & \tabincell{c}{STARK \\ \cite{stark}} & \tabincell{c}{SBT \\ \cite{SBT}}  & \tabincell{c}{Mixformer \\ \cite{mixformer}} & \tabincell{c}{SeqTrack \\ \cite{seqtrack}} & \tabincell{c}{ODTrack \\ \cite{odtrack}} & \tabincell{c}{SSTrack-256 \\ \cite{SSTrack}} & \tabincell{c}{SSTrack-384 \\ \cite{SSTrack}} & \textbf{{\tracker}-256} & \textbf{{\tracker}-384} \\
\midrule
EAO$(\uparrow)$ & 0.321 & 0.430 & 0.439 & 0.482 & 0.491 & 0.505 & 0.515 & 0.535 & 0.522 & 0.581 & {0.458} & \color{blue}{0.503} & {0.483} & \color{red}{0.534} \\
Accuracy$(\uparrow)$ & 0.624 & 0.693 & 0.699 & 0.754 & 0.685 & 0.759 & 0.752 & 0.761 & - & 0.764 & {0.664} & \color{blue}{0.754} & {0.686} & \color{red}{0.759} \\
Robustness$(\uparrow)$ & 0.648 & 0.754 & 0.769 & 0.777 & 0.842 & 0.819 & 0.825 & 0.854 & - & 0.877 & {0.839} & {0.816} & \color{blue}{0.844} & \color{red}{0.846} \\
\bottomrule
\end{tabular} }
\end{table*}

\textbf{TrackingNet.}
TrackingNet \cite{trackingnet} is a large-scale tracking benchmark with extensive manual bounding box annotations, offering a testset with 511 video sequences. As shown in Tab.\ref{tab:results}, our tracker achieves excellent results in short-term tracking scenarios. For example, compared to the self-supervised method TADS, SSTrack surpasses it by 14.8\% in AUC score. Additionally, compared to the state-of-the-art supervised tracker ARTrackV2 \cite{Artrackv2}, SSTrack is only 3.5\% lower in normalized precision, significantly enhancing the potential of self-supervised tracking algorithms. By implementing dynamic temporal association of target instance, our {\tracker}-256 gets a success score (AUC) of 81.4\%, a normalized precision score ($P_{Norm}$) of 87.1\%, and a precision score (P) of 79.1\%.
These results validate our core motivation: by repurposing the model's rich output representations, fine-grained prompt cues about the target instance can be effectively mined to enhance self-supervised tracking performance.

\textbf{LaSOT$_{\rm{ext}}$.}
Compared to the LaSOT dataset, LaSOT$_{\rm{ext}}$ \cite{lasot-ext} is a large-scale tracking dataset that includes more challenging video sequences. Most state-of-the-art supervised trackers are evaluated on this benchmark to verify their accuracy and robustness. As shown in Tab.\ref{tab:results}, our self-supervised tracking framework achieves competitive results and significantly narrows the performance gap with fully supervised methods. Specifically, our {\tracker}-384 gets a AUC score of 49.1\%, a normalized precision score ($P_{Norm}$) of 61.3\%, and a precision score (P) of 55.8\%. Compared to SSTrack, {\tracker} equipped with the self-prompting evolution module achieves additional performance gains of 0.6\% in AUC, 0.4\% in $P_{norm}$, and 1.3\% in precision. These results demonstrate the effectiveness of our proposed method, achieving good tracking performance even with limited annotations.

\begin{table}[t]
\centering
\caption{Comparison with SOTA methods on OTB100 \cite{OTB2015}, UAV123 \cite{uav123}, and VOT2018 \cite{VOT2018} datasets. The best two results are shown in {\color{red}red} and {\color{blue}blue} fonts.
}
\label{tab:OTB}
\setlength{\tabcolsep}{0.5mm}{
\begin{tabular}{l|ccc}
\toprule
Method & OTB(AUC) & UAV(AUC) & VOT18(Acc) \\
\midrule
\multicolumn{4}{l}{\multirow{1}{*}{\textit{Fully Supervised Tracking}}} \\
ATOM \cite{atom} & 67.1 & 64.3 & 0.590 \\
Ocean \cite{Ocean} & 68.4 & - & 0.592 \\
DiMP \cite{DiMP50} & 68.4 & 65.3 & 0.597 \\
TransT \cite{transt} & 69.4 & 69.1 & - \\
TrDiMP \cite{trdimp} & 67.5 & 67.5 & - \\
MixFormer \cite{mixformer} & 70.0 & 70.4 & - \\
HIPTrack \cite{HIPTrack} & 71.0 & 70.5 & - \\
LoRAT-B378 \cite{lorat} & 71.7 & 71.9 & - \\
DreamTrack \cite{dreamtrack} & 72.0 & 72.5 & - \\
\midrule
\multicolumn{4}{l}{\multirow{1}{*}{\textit{Self Supervised Tracking}}} \\
S$^2$SiamFC \cite{s2siamfc} & - & - & 0.463 \\
CycleSiam \cite{cycleSiam} & - & - & 0.562 \\
self-SDCT \cite{SDCT} & 63.8 & 50.1 & - \\
TADS \cite{TADS} & 65.3 & 55.2 & - \\
SSTrack-256 \cite{SSTrack} & 67.9 & 65.5 & 0.587 \\
SSTrack-384 \cite{SSTrack} & \color{blue}70.5 & \color{blue}66.1 & \color{blue}0.630 \\
\textbf{\tracker-256} & {69.7} & {65.0} & {0.587} \\
\textbf{\tracker-384} & \color{red}{70.9} & \color{red}{66.3} & \color{red}{0.631} \\
\bottomrule
\end{tabular}
}
\end{table}

\begin{table}[t]
\centering
\caption{Comparison with SOTA methods on VastTrack \cite{vasttrack} dataset. The best two results are shown in {\color{red}red} and {\color{blue}blue} fonts.
}
\label{tab:vasttrack}
\setlength{\tabcolsep}{3.5mm}{
\begin{tabular}{l|ccc}
\toprule
Method & AUC & P$_\text{Norm}$ & P \\
\midrule
\multicolumn{4}{l}{\multirow{1}{*}{\textit{Fully Supervised Tracking}}} \\
SiamFC \cite{SiamFC} & 8.5 & 2.9 & 4.3 \\
ATOM \cite{atom} & 17.2 & 14.1 & 10.2 \\
Ocean \cite{Ocean} & 27.5 & 30.2 & 24.6 \\
DiMP \cite{DiMP50} & 29.9 & 31.7 & 25.7 \\
TransT \cite{transt} & 29.9 & 31.4 & 25.4 \\
STARK \cite{stark} & 33.4 & 34.3 & 30.8 \\
OSTrack \cite{ostrack} & 33.6 & 34.5 & 31.5 \\
DropMAE \cite{DropMAE} & 37.5 & 39.7 & 36.5 \\
Mixformer \cite{mixformer} & 39.5 & 42.4 & 39.8 \\
SeqTrack \cite{seqtrack} & 39.6 & 42.9 & 40.2 \\
\midrule
\multicolumn{4}{l}{\multirow{1}{*}{\textit{Self Supervised Tracking}}} \\
SSTrack-256 \cite{SSTrack} & 32.4 & 42.2 & 31.2 \\
SSTrack-384 \cite{SSTrack} & \color{blue}34.7 & \color{blue}44.3 & \color{blue}34.0 \\
\textbf{\tracker-256} & 34.1 & 43.2 & 32.2 \\
\textbf{\tracker-384} & \color{red}{36.1} & \color{red}{45.8} & \color{red}{35.5} \\
\bottomrule
\end{tabular}
}
\end{table}

\textbf{TNL2K.}
TNL2K \cite{tnl2k} is a large-scale vision-language tracking dataset that includes 1300 training video sequences and 700 testing sequences. As shown in Tab.\ref{tab:tnl2k}, our SSTrack and {\tracker} achieve competitive performance compared to most state-of-the-art fully-supervised trackers. For instance, our {\tracker}-384 achieves comparable result of 54.1\% and 56.5\% in terms of AUC and precision score. Notably, it trails the fully-supervised method AQATrack by only 5.2\% in AUC and 5.8\% in precision. This narrowing of the performance gap demonstrates the significant real-world potential of the proposed self-supervised model.

\textbf{VOT2020.}
VOT2020 \cite{VOT2020} contains 60 challenging sequences, and it uses binary segmentation masks as the groundtruth. We use Alpha-Refine \cite{Alpha-Refine} as a segmentation model for {\tracker} to predict segmentation masks. As shown in Tab.\ref{tab:vot2020}, our SSTrack-384 and {\tracker}-384 achieve the best self-supervised tracking results with EAO of 50.3\% and 53.4\% on mask evaluations. These performance gains indicate that {\tracker}, equipped with the self-prompting evolution module, effectively leverages fine-grained target tokens to improve the generalization capability of the model in unlabeled tracking scenarios.

\textbf{OTB100, UAV123, and VOT2018.}
OTB100 \cite{OTB2015}, UAV123 \cite{uav123}, and VOT2018 \cite{VOT2018} are widely-used visual tracking datasets, comprising a variety of video sequences that pose challenges such as occlusion, lighting variations, motion changes, and camera motion. As indicated in Tab.\ref{tab:OTB}, on the OTB100 dataset, compared with the previous fully supervised MixFormer tracker, our self-supervised SSTrack++-384 model achieves a 0.9\% improvement in AUC score. However, there remains a 1.1\% gap compared to the latest fully supervised DreamTrack tracker. These results demonstrate the effectiveness of our proposed self-supervised framework in narrowing the performance gap between self-supervised and fully supervised approaches.

On the UAV23 dataset, our SSTrack++-256 method surpasses the strong self-supervised tracker TADS by 9.8\% in AUC, validating the effectiveness of the proposed self-prompt evolution module in significantly enhancing performance in UAV tracking scenarios.

On the VOT2018 dataset, compared with the CycleSiam self-supervised tracker, our SSTrack++-256 model improves the Acc score by 2.5\%, further confirming that the proposed weak-to-strong self-supervised training framework effectively enhances robustness in challenging scenarios.

\begin{table}[t]
  \centering
  \caption{Ablation studies of our SSTrack variants in GOT-10k.}
  \label{tab:module}
  \setlength{\tabcolsep}{0.9mm}{
  \begin{tabular}{cc|ccc}
    \toprule
    \# & Method & AO & SR$_{0.5}$ & SR$_{0.75}$ \\
    \midrule
    1 & Baseline & 41.8 & 45.2 & 14.3 \\
    \midrule
    2 & Decoupled training w/o context & 68.6 & 78.1 & 60.1 \\
    3 & + Spatio-temporal context & 70.8 & 82.2 & 64.6 \\
    4 & + Instance contrastive loss & 72.4 & 83.6 & 66.2 \\
    \bottomrule
  \end{tabular}
  }
\end{table}

\begin{table}[t]
  \centering
  \caption{Comparison of different pre-trained models.}
  \label{tab:pretrain}
  \begin{tabular}{cc|ccc}
    \toprule
    \# & Pre-trained Model & AO & SR$_{0.5}$ & SR$_{0.75}$ \\
    \midrule
    1 & MAE & 62.7 & 71.1 & 54.1 \\
    2 & DropMAE & 68.6 & 78.1 & 60.1 \\
    \bottomrule
  \end{tabular}
\end{table}

\textbf{VastTrack.}
we evaluate it on the VastTrack \cite{vasttrack} dataset without using it for training. As shown in Tab.\ref{tab:vasttrack}, on diverse unseen scenarios, the fully supervised OSTrack model achieves 33.6\%, 34.5\%, and 31.5\% in AUC, P$_\text{Norm}$, and P scores, respectively. In comparison, our {\tracker}-256 surpasses it by 0.5\%, 8.7\%, and 0.7\% on these metrics. These results demonstrate that the proposed {\tracker} equipped with the SPE module can effectively adapt to diverse tracking scenarios, indicating strong generalization ability. Notably, compared with all fully supervised methods, our {\tracker}-384 achieves the best performance in P$_\text{Norm}$ metric, suggesting strong robustness to target scale variations and superior multi-scale generalization capability.

\subsection{Ablation Study} \label{study}

To explore the role of the various components in the proposed self-supervised tracking frameworks, we conduct comprehensive ablation studies on the GOT10K \cite{got10k} and LaSOT \cite{lasot} benchmarks.

\textbf{Importance of decoupled spatio-temporal consistency learning strategy.}
As shown in Tab.\ref{tab:module}, \textit{baseline} represents a self-supervised model based on contrastive learning. When we introduce our decoupled training framework, its performance significantly improves, achieving an increase of 26.8\% in AO score. By incorporating spatio-temporal context into our self-supervised framework, the tracking performance improves by an additional 2.2\% in AO score. These results indicate that the decoupled training framework effectively learns instance correspondences across various tracking scenarios, playing a crucial role in our self-supervised tracking framework.

\begin{table}[t]
  \centering
  \caption{Comparison of different data augmentations.}
  \label{tab:aug}
  \begin{tabular}{cccc|ccc}
    \toprule
    \# & Shear & Blur & LSJ & AO & SR$_{0.5}$ & SR$_{0.75}$ \\
    \midrule
    1 & \checkmark & \checkmark & \checkmark & 70.3 & 81.2 & 63.3 \\
    2 & \ding{55} &  &  & 69.4 & 80.5 & 62.4 \\
    3 &  & \ding{55} &  & 69.3 & 79.8 & 61.3 \\
    4 &  &  & \ding{55} & 69.0 & 80.0 & 61.6 \\
    5 &  &  &  & 68.6 & 78.1 & 60.1 \\
    \bottomrule
  \end{tabular}
\end{table}

\begin{table}[t]
  \centering
  \caption{Ablation studies of our {\tracker} variants on LaSOT dataset.}
  \label{tab:modulev2}
  \setlength{\tabcolsep}{1.3mm}{
  \begin{tabular}{cc|ccc}
    \toprule
    \# & Method & AUC & P$_{\text{Norm}}$ & P \\
    \midrule
    1 & {\tracker} & 66.3 & 76.6 & 70.8 \\
    2 & - Weak-to-Strong Framework & 65.9 & 76.0 & 70.2 \\
    3 & - Self-Prompting Evolution & 65.3 & 76.1 & 69.7 \\
    \midrule
    4 & + Query Association & 65.0 & 75.8 & 69.5 \\
    \bottomrule
  \end{tabular}
  }
\end{table}

\textbf{Importance of instance contrastive loss.}
As shown in the fourth row of Tab.\ref{tab:module}, by adding the instance contrastive loss to our self-supervised framework, our tracker improves by 1.6\%, 1.4\%, and 1.6\% in AO, SR$_{0.5}$, and SR$_{0.75}$ metrics, respectively. This result demonstrates that the instance contrastive loss effectively enhances the discriminability of instance representations in self-supervised learning, thereby improving the tracking performance of our model.

\begin{figure}[t]
      \centering
      \includegraphics[width=0.85\columnwidth]{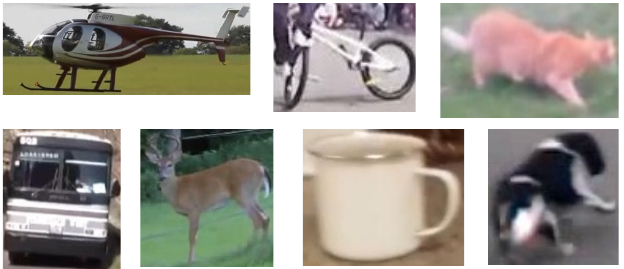}
       \caption{Target instances sampled by the instance contrastive learning method.}
       \label{fig:instance}
    \end{figure}

Furthermore, as illustrated in Fig.\ref{fig:instance}, we visualize the target instances extracted during training to demonstrate that the proposed strategy can accurately capture true target instances in generic scenes.
In addition, in certain highly challenging scenarios (e.g., similar-object), accurately identifying the true target instance remains difficult to fully resolve. This continues to be an open and challenging problem for the tracking community. A potential direction is to draw inspiration from re-identification techniques by designing fine-grained target identity association mechanisms, enabling the model to better distinguish and consistently track the true target in ambiguous scenarios.

   \begin{figure}[t]
      \centering
      \includegraphics[width=0.9\linewidth]{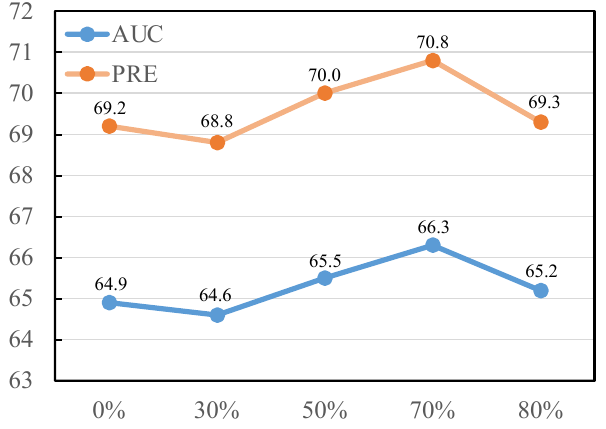}
       \caption{Comparison of different filtering ratios $\lambda$ on LaSOT dataset.}
       \label{fig:ratio}
    \end{figure}

   \begin{figure*}[t]
      \centering
      \includegraphics[width=1\linewidth]{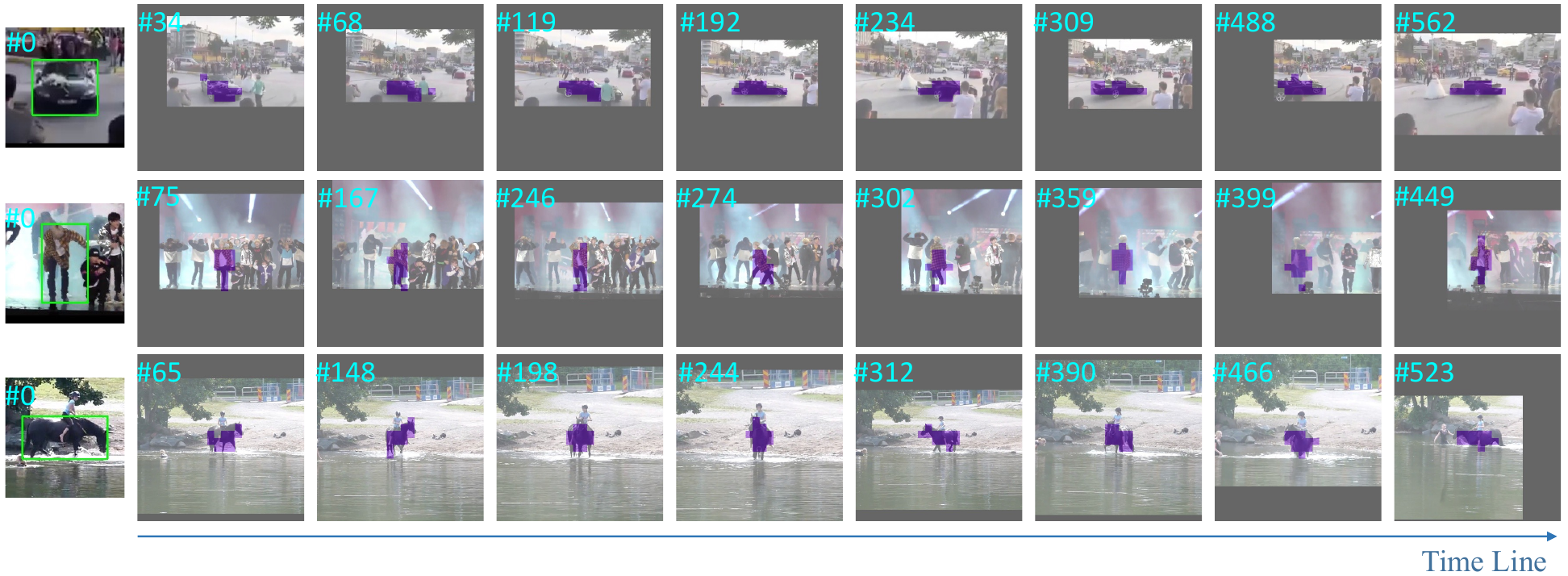}
       \caption{Visualization of the target self-prompts evolved on TrackingNet benchmark.
       }
       \label{fig:prompt_tokens}
    \end{figure*}

\textbf{Effect of pre-trained model.}
We conduct experiments in Tab.\ref{tab:pretrain} to validate the impact of different pre-trained models on our model. We observe that by replacing the MAE \cite{MAE} pre-trained model with DropMAE, our model achieves a performance gain of 5.9\% in AO score. This demonstrates that DropMAE, with temporal correspondence learning, is more suitable for self-supervised tracking task and more effectively learns target correspondences from unlabeled data.

\textbf{Effect of data augmentations.} 
Due to the absence of annotations, self-supervised algorithms struggle to achieve sufficient training. To address this issue, we incorporate various data augmentation methods such as shear, blur, and LSJ \cite{LSJ}. As shown in Tab.\ref{tab:aug}, these augmentation methods enhance our model's ability to learn target variations from different views by increasing data diversity. In other words, appropriate data augmentation methods improve our model's robustness.

\begin{table}[t]
  \centering
  \caption{Comparison of different self-prompt lengths on LaSOT bwnchmark.}
  \begin{tabular}{c|c|ccc}
    \toprule
    \# & Prompt Number & AUC & P$_\text{Norm}$ & P \\
    \midrule
    1 & {4} & 65.4 & 76.0 & 69.4 \\
    2 & {8} & 66.3 & 76.6 & 70.8 \\
    3 & {16} & 65.9 & 76.6 & 70.7 \\
    4 & {24} & 64.3 & 75.4 & 68.7 \\
    5 & {32} & 61.5 & 70.1 & 62.1 \\
    \bottomrule
  \end{tabular}
  \label{tab:prompt_number}
\end{table}

\textbf{Importance of weak-to-strong training framework.}
As shown in Tab.\ref{tab:modulev2} (\#2), when the weak-to-strong training framework is removed, the AUC score of our self-supervised tracker drops by 0.4\%, indicating that the absence of effective strong–weak supervisory signals for large amounts of unlabeled video frames severely hinders model learning. These results demonstrate that the proposed strong–weak visual cues provide effective guidance for training self-supervised trackers, enabling reliable acquisition of tracking knowledge and ensuring stable self-supervised model optimization.

   \begin{figure}[t]
      \centering
      \includegraphics[width=1\linewidth]{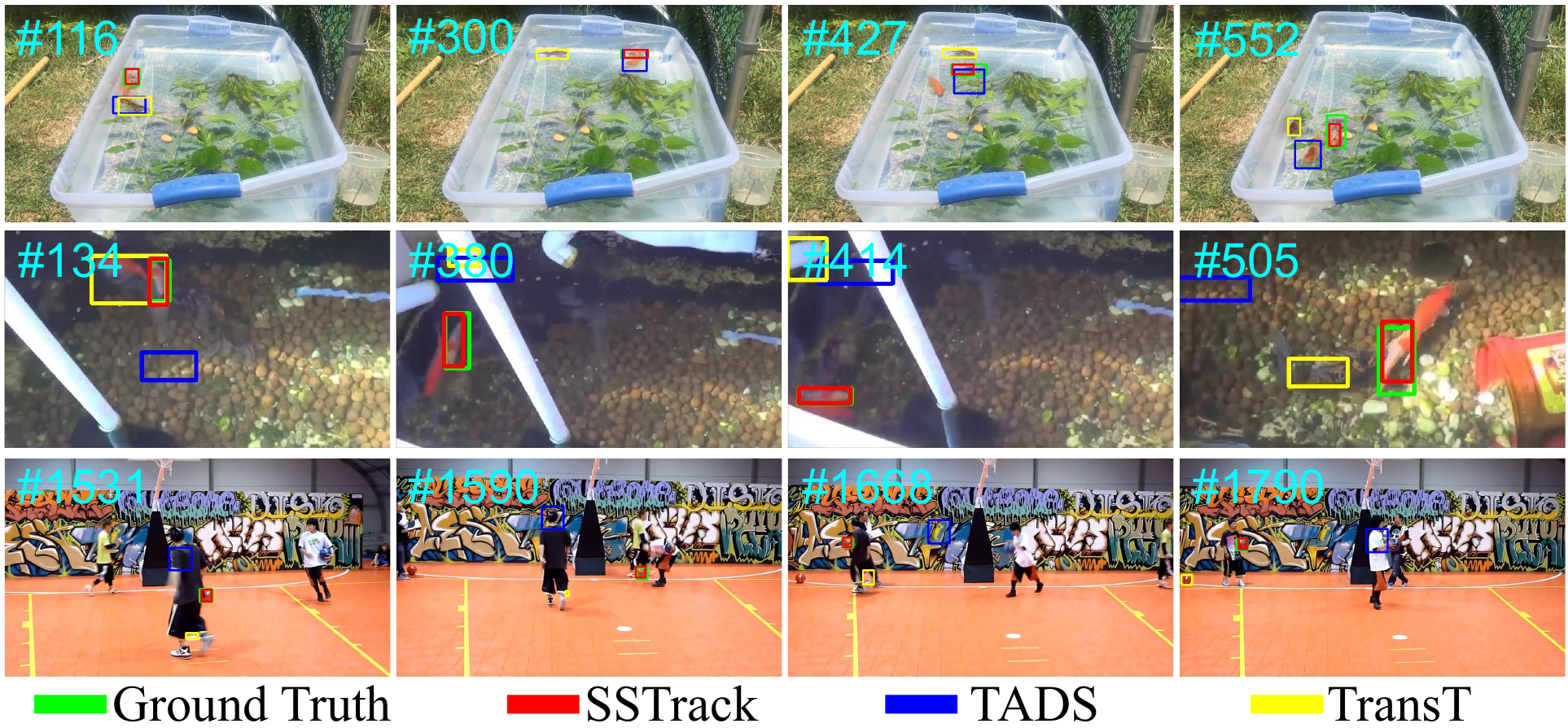}
       \caption{Visualization of our tracker with self- and fully- supervised trackers on LaSOT benchmark.
       }
       \label{fig:visual}
    \end{figure}

\textbf{Importance of Self-Prompting Evolution Module.}
As shown in Tab.\ref{tab:modulev2} (\#3), compared with the complete self-supervised tracker, {\tracker} without the Self-Prompting Evolution (SPE) module suffers AUC and precision drops of 1\% and 1.1\%, respectively. This performance degradation highlights the necessity of incorporating self-prompt evolution into the self-supervised tracking framework, as it enables continuous collection and refinement of target-specific self-prompts that are most relevant to the current tracking scenario. Such adaptive self-prompts provide richer target-aware supervisory signals in unlabeled settings, thereby improving self-supervised tracking performance.

Furthermore, we compare the proposed SPE module with the conventional query-based association strategy. As shown in Tab.\ref{tab:modulev2} (\#4), our SPE module can accurately perceive the target’s spatial location and fine-grained appearance details in unlabeled videos, whereas traditional query association methods fail to achieve this. These results further validate the effectiveness of the SPE module for self-supervised tracking.

\textbf{Comparison of the number of self-prompts.}
The number of self-prompts also has a noticeable impact on the performance of our self-supervised tracker. To investigate this effect, we conduct a comparative study by varying the number of self-prompts. As shown in the Tab.\ref{tab:prompt_number}, increasing the number of self-prompts initially improves tracking performance, but further increases lead to a performance drop. Our analysis reveals that, under fixed input resolution settings, an excessive number of self-prompts is likely to sample noisy background tokens, which in turn degrades model performance. Therefore, selecting an appropriate number of self-prompts is crucial for maintaining stable and reliable performance.

\textbf{Comparison of different filtering ratios $\lambda$.}
As illustrated in the Fig.\ref{fig:ratio}, we evaluate the impact of background noise tokens on self-supervised tracking by varying the token filtering ratio $\lambda$. In unlabeled settings, increasing the filtering ratio gradually improves the tracking performance, indicating the effectiveness of suppressing background noise. However, excessively aggressive filtering may inadvertently remove informative target tokens, leading to performance degradation. Overall, a filtering ratio of 70\% yields the best trade-off, allowing {\tracker} to discard redundant background tokens while preserving accurate target regions.

\begin{table}[t]
\footnotesize
\centering
\caption{{Comparison of model parameters, FLOPs, and inference speed.}}
\label{tab:param}
\setlength{\tabcolsep}{0.4mm}{
\begin{tabular}{l|cccccc}
\toprule
Tracker & Type & Resolution & Params & FLOPs & Speed  & Device \\
\midrule
SeqTrack & ViT-B & $384\times384$ & 89M & 148G & 21$fps$ & A100 \\
AQATrack & HiViT-B & $384\times384$ & 72M & 58G & 57$fps$ & A100 \\
{\tracker} & ViT-B & $384\times384$ & 92M & 73G & 59$fps$ & A100 \\
\bottomrule
\end{tabular} }
\end{table}

\subsection{Qualitative Analysis}

\textbf{Visualization.}
To intuitively verify whether self-prompts are progressively selected and evolved toward accurate target representations over time in unlabeled tracking scenarios, we conduct comprehensive qualitative visualizations, as shown in Fig.\ref{fig:prompt_tokens}. Specifically, in the first video sequence, as the vehicle (target) moves forward, it becomes partially occluded by pedestrians. Despite this occlusion, the evolved self-prompts remain accurately localized within the target region while effectively avoiding pedestrian areas. These observations demonstrate that our self-prompt evolution module retains strong target-awareness even in unlabeled settings, thereby enabling the {\tracker} to achieve robust self-supervised tracking across diverse and challenging scenarios.

Furthermore, we conduct qualitative experiments to visually demonstrate the effectiveness of the proposed framework. Fig.\ref{fig:visual} presents a visual comparison of our {\tracker} with the advanced self-supervised tracker TADS and the fully-supervised tracker TransT. By effectively learning the spatio-temporal context of object in a self-supervised manner and improving the instance features discriminability through contrastive learning, our model performs exceptionally well in various complex scenarios, even rivaling fully supervised tracking algorithms.

\textbf{Speed, FLOPs and Params Analysis.}
On the other hand, we analyze the parameters, FLOPs, and inference speed of different models. As shown in Tab.\ref{tab:param}, our {\tracker} runs at 59 \textit{fps} on an A100 GPU. Compared to SeqTrack \cite{seqtrack} and AQATrack \cite{AQATrack}, we achieve faster inference speed.

\section{Conclusion}\label{conclusion}

In this work, we have proposed a self-supervised tracking framework named {\tracker}, aimed at eliminating the reliance on costly box annotations. Specifically, under a weak-to-strong self-supervised training paradigm, we have introduced a simple yet efficient decoupled spatio-temporal consistency learning strategy to learn rich target appearance and motion information across timestamps. Furthermore, we have designed a self-prompt evolution module that jointly mines target appearance evolution at both the feature and decision levels, enabling more robust adaptation to complex and diverse unlabeled tracking scenarios. Finally, we have proposed an instance contrastive loss function to learn instance-level correspondences from a multi-view perspective, providing reliable instance supervision without any labels. Extensive experiments have demonstrated the superiority of our method. We hope this work will further inspire research into self-supervised tracking algorithms.

\textbf{\textit{Limitation.}}
This work leverages the powerful advantages of the decoupled spatio-temporal consistency learning strategy and self-prompting evolution module to design a novel self-supervised tracking algorithm. Despite achieving remarkable results, we observe that the performance of backward tracking somewhat depends on the localization accuracy of forward tracking. Thus, improving the accuracy of forward tracking could further enhance the performance of self-supervised tracking, potentially narrowing the gap with fully supervised tracking algorithms even more.

\backmatter





\bmhead{Acknowledgements}
This work is supported by the Project of Guangxi Scienceand Technology (No.2024GXNSFGA010001, and GuiKeFN2504240017), the National Natural Science Foundation of China (No.U23A20383, 62472109, and 62466051), the Guangxi "Young Bagui Scholar" Teams for Innovation and Research Project, and the Research Project of Guangxi Normal University (No.2026DFO001).








\bibliographystyle{plain}
\bibliography{reference} 

\newpage

\section{Appendix}

\textbf{Additional Backgrounds.} 
With the advancement of deep learning techniques \cite{xiao2026prototype,zhou2026comem,xiaopath,zhang2026pi,xiao2026reversible,LEADer,OFFSET,HABIT,ENCODER,ReTrack,STABLE,ConeSep,gao2023UTrack,he2023camouflaged,he2025segment,lu2023tf,lu2024mace,peng2024lightweight} and the potential to eliminate the need for large-scale labeled data, self-supervised tracking has attracted increasing attention from researchers. Taking advantage of intrinsic correlations in unlabeled video data, such as temporal consistency, self-supervised tracking has shown promising results in relatively simple tracking scenarios. However, in long-term complex unlabeled tracking settings, it remains a significant challenge to capture cross-frame motion patterns and to learn robust target representations.

\textbf{Evaluation Metrics.} The tracking performance is evaluated using the toolkit corresponding to the dataset. We follow the evaluation protocol of published datasets and employ three metrics to ensure a fair comparison across various tracking methods, including success score (AUC), normalized precision score (P$_{\rm{Norm}}$), and precision score (P).

\begin{table}[h]
  \centering
  \caption{Comparison of different loss weights on the LaSOT benchmark.}
  \label{tab:loss}
  \setlength{\tabcolsep}{3mm}{
  \begin{tabular}{cc|ccc}
    \toprule
    \# & Weights & AUC & P$_\text{Norm}$ & P \\
    \midrule
    1 & 1:1:1 & 66.3 & 76.6 & 70.8 \\
    2 & 0.5:1:1 & 65.3 & 75.6 & 69.2 \\
    3 & 1:0.5:1 & 65.7 & 76.2 & 70.1 \\
    4 & 1:1:0.5 & 65.9	& 76.7 & 70.0 \\
    \bottomrule
  \end{tabular}
  }
\end{table}

\textbf{Effect of different loss terms and weights.}
We conduct experiments in Tab.\ref{tab:loss} to validate the impact of different loss weights ($\omega_1$ : $\omega_2$ : $\omega_3$) on our model. 
As shown in Tab.\ref{tab:loss}, varying the weighting ratio leads to different self-supervised performance. In particular, reducing the weight of any individual loss term consistently results in a slight performance drop, indicating that all three losses play indispensable roles during training. Notably, when using an equal weighting scheme ($\omega_1 : \omega_2 : \omega_3 = 1:1:1$), the model achieves strong performance, demonstrating that {\tracker} is not sensitive to loss weighting and maintains stable self-supervised tracking performance.
Furthermore, we instead conducted experiments by removing L$_\text{u}$ and L$_\text{mse}$, respectively. As shown in the Tab.\ref{tab:loss_term}, the results confirm that each component contributes to the overall performance to a certain extent.

\begin{table}[h]
  \centering
  \caption{Comparison of different loss terms on the LaSOT benchmark.}
  \label{tab:loss_term}
  \setlength{\tabcolsep}{3mm}{
  \begin{tabular}{cc|ccc}
    \toprule
    \# & Methods & AUC & P$_\text{Norm}$ & P \\
    \midrule
    1 & {\tracker} & 66.3 & 76.6 & 70.8 \\
    2 & - L$_\text{u}$ & 65.2 & 75.9 & 69.0 \\
    3 &  - L$_\text{mse}$ & 65.6 & 76.6 & 69.7 \\
    \bottomrule
  \end{tabular}
  }
\end{table}

\end{document}